\documentclass{article} 
\usepackage{iclr2024_conference,times}

\usepackage{amsmath,amsfonts,bm}

\def\eqref#1{equation~\ref{#1}}

\def\1{\bm{1}}

\DeclareMathAlphabet{\mathsfit}{\encodingdefault}{\sfdefault}{m}{sl}
\SetMathAlphabet{\mathsfit}{bold}{\encodingdefault}{\sfdefault}{bx}{n}

\usepackage{hyperref}
\usepackage{url}
\usepackage{graphicx}
\usepackage{caption}
\usepackage{subcaption}
\usepackage{bm}
\usepackage{amsmath,amssymb}
\usepackage{booktabs}
\usepackage{multirow}
\usepackage[flushleft]{threeparttable}
\usepackage{makecell}
\usepackage{xspace}
\usepackage{color}
\usepackage{wrapfig}
\usepackage{bbding}
\usepackage{ifthen}
\usepackage{verbatim}
\usepackage{pifont}
\usepackage{caption}

\newcommand{\eg}{\emph{e.g.,}\xspace}

\usepackage{siunitx}
\usepackage{dcolumn}

\usepackage{graphicx}

\newcommand{\dacheng}[1]{{\color{black} #1}}

\newboolean{showtodos}
\setboolean{showtodos}{true}

\newcommand{\todo}[1]{
  \ifthenelse{\boolean{showtodos}}{
    \textcolor{red}{\textbf{TODO:} #1}
  }{}
}
\newcommand{\zigzag}{\textsc{ZigZag-RingAttn}\xspace}
\definecolor{DLTODO}{RGB}{0,0,0}
\definecolor{YELLOW}{rgb}{0.83, 0.61, 0.18}
\definecolor{nvgreen}{RGB}{118, 185, 0}

\usepackage{graphicx}
\usepackage{caption}
\usepackage{bm}
\usepackage{booktabs}
\usepackage{inconsolata}
\usepackage{listings}
\usepackage{graphicx}

\usepackage[most]{tcolorbox}

\tcbuselibrary{xparse,skins,breakable,listings}

\usepackage{cleveref}

\tcbset{%
    basestyle/.style={enhanced,breakable,
        fonttitle=\scshape,
        coltitle=white},
    defstyle/.style={basestyle},
    theostyle/.style={basestyle},
}
\DeclareTColorBox[auto counter,crefname={example}{examples}]{example}{ o O{} t\label g }{%
    enhanced,
    listing side text,defstyle,IfValueTF={#1}{title={Conclusion~\thetcbcounter\ (#1)}}{title=Conclusion}, IfBooleanTF={#3}{label=#4}{},#2}

\title{LongVILA: Scaling Long-Context Visual Language Models for Long Videos}

\author{
\begin{minipage}[t]{\textwidth}
\centering
Yukang Chen$^1$\thanks{Algorithm Lead. $^{\dagger}$ System Lead. The first four authors have equal contributions.} \qquad Fuzhao Xue$^1$\footnotemark[1] \qquad Dacheng Li$^{1,3 \dagger}$ \qquad Qinghao Hu$^{2}$\footnotemark[2] \\[0.1cm]
Ligeng Zhu$^1$ \qquad Xiuyu Li$^{1,3}$ \qquad
Yunhao Fang$^1$ \qquad Haotian Tang$^{1,2}$ \qquad Shang Yang$^{1,2}$ \\[0.1cm]
\qquad Zhijian Liu$^1$ \qquad Ethan He$^1$ \qquad Hongxu Yin$^1$ \qquad Pavlo Molchanov$^1$ \qquad Jan Kautz$^1$ \qquad\\[0.1cm]
Linxi Fan$^1$ \qquad Yuke Zhu$^{1,4}$ \qquad Yao Lu$^1$ \qquad Song Han$^{1,2}$ \\[0.3cm]
$^1$NVIDIA \qquad $^2$MIT \qquad  $^3$UC Berkeley \qquad $^4$UT Austin \\
\end{minipage}
}

\iclrfinalcopy 
\begin{document}
\maketitle
\begin{abstract}
Long-context capability is critical for multi-modal foundation models, especially for long video understanding. We introduce LongVILA, a full-stack solution for long-context visual-language models {by co-designing the algorithm and system. For model training, we upgrade existing VLMs to support long video understanding by incorporating two additional stages, {\em i.e.}, long context extension and long video supervised fine-tuning.
However, training on long video is computationally and memory intensive. We introduce the long-context Multi-Modal Sequence Parallelism (MM-SP) system that efficiently parallelizes long video training and inference, enabling 2M context length training on 256 GPUs without any gradient checkpointing.
{
LongVILA efficiently extends the number of video frames of VILA from 8 to 2048, achieving 99.8\% accuracy in 6,000-frame (more than 1 million tokens) video needle-in-a-haystack. LongVILA-7B demonstrates strong accuracy on 9 popular video benchmarks, \eg 65.1\% VideoMME with subtitle.} Besides, MM-SP is $2.1 \times$ - $5.7 \times$ faster than ring style sequence parallelism and $1.1 \times$ - $1.4 \times$ faster than Megatron with a hybrid context and tensor parallelism. Moreover, it seamlessly integrates with Hugging Face Transformers. Our code and models are available at \href{https://github.com/NVlabs/VILA/tree/main/longvila}{github.com/NVlabs/VILA/longvila}.
}
\end{abstract}

\section{Introduction}

Integrating multi-modal understanding with long-context capability is important.
A foundation model supporting more modalities can take more flexible input signals so that people can interact with the model in more diverse manners, \eg GPT-4o-like multi-modal chatbot, multi-modal web agent~\citep{koh2024visualwebarena}, and real-world robotics foundation model~\citep{brohan2022rt,brohan2023rt,padalkar2023open}. Longer context enables models to process more information, \eg long documents, repo-level codebase, and hour-length video, which similarly provides required features to more real-world applications. 

While some works have enabled long-context Vision-Language Models (VLMs)~\citep{lin2023vila, longvlm}, they employ simplified approaches rather than offering a comprehensive solution. For instance, LongVA~\citep{zhang2024longva} relies on long-context LLMs and trains models on short-context data. LongVLM~\citep{longvlm} utilizes token compression to circumvent context extension. These approaches sidestep more challenging issues, such as the development of a robust long-context multi-modal training framework and corresponding dataset design. 

A full-stack design is crucial for long-context Vision-Language Models (VLMs). Training large models is typically a complex, systematic endeavor that demands both data engineering~\citep{betker2023improving,ouyang2022training,zhou2024lima} and system-software co-design~\citep{lepikhin2020gshard,chowdhery2023palm,shoeybi2019megatron,brown2020language,dehghani2023scaling}. Unlike text-only LLMs, VLMs (\eg LLaVA~\citep{liu2023llava}) often require distinct model architectures and flexible distributed training strategies. Additionally, long-context modeling necessitates not only long-context data to fully utilize the model's capabilities~\citep{fu2024data,chen2023longlora} but also infrastructure capable of supporting memory-intensive long-context training~\citep{li2021sequence,jacobs2023deepspeed,li2023lightseq}. Therefore, a full-stack design, {encompassing training pipeline and system,} is indispensable for long-context VLMs.

\begin{figure}[t]
    \centering
    \includegraphics[width=\linewidth]{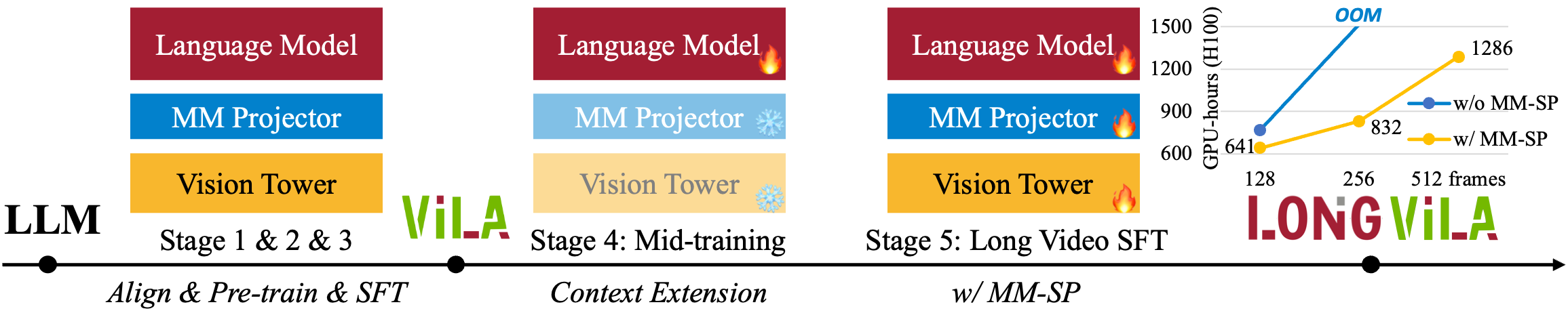}
    \caption{{The LongVILA training pipeline. In Stages 1 through 3, the process starts with alignment, pre-training, and supervised fine-tuning. In Stage 4, the model undergoes mid-training context extension. Finally, in Stage 5, the model is fine-tuned for long video understanding with Multi-Modal Sequence Parallelism~(MM-SP).}}
    \label{fig:training-pipeline}
\end{figure}

\begin{figure}[t]
    \centering
    \includegraphics[width=\linewidth]{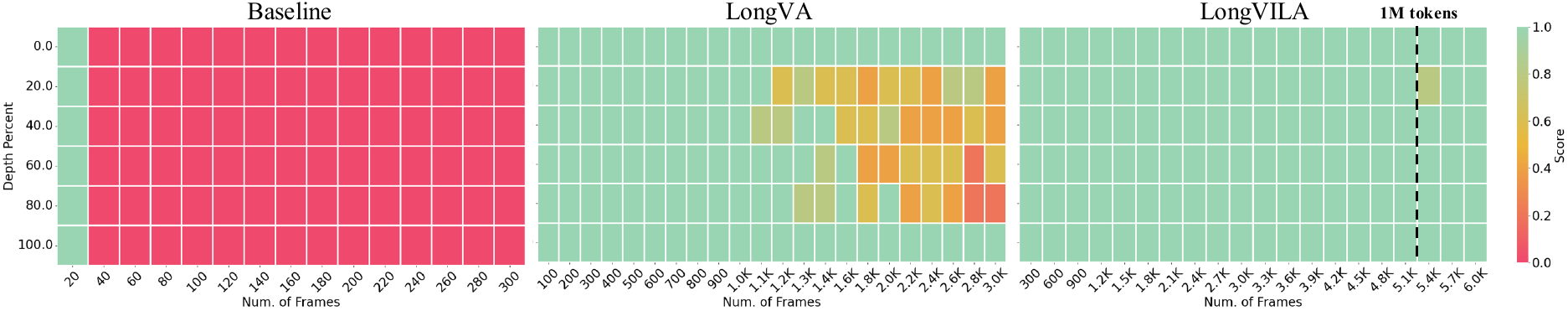}
    \caption{{Comparison of Needle in the Long Video Haystack Experiment. The 32-frame baseline model (left) can not retrieve right needles after 32 frames. LongVA (middle) achieves 87.6\% accuracy in 3,000 frames. In contrast, the LongVILA model (right),  trained on 2048 frames, presents 99.8\% accuracy on 6,000 frames (more than 1 million context length).}}
    \label{fig:needle-in-a-haystack}
\end{figure}

\begin{figure}[t]
  \begin{minipage}[b]{0.46\textwidth}
    \centering
    \includegraphics[width=\textwidth]{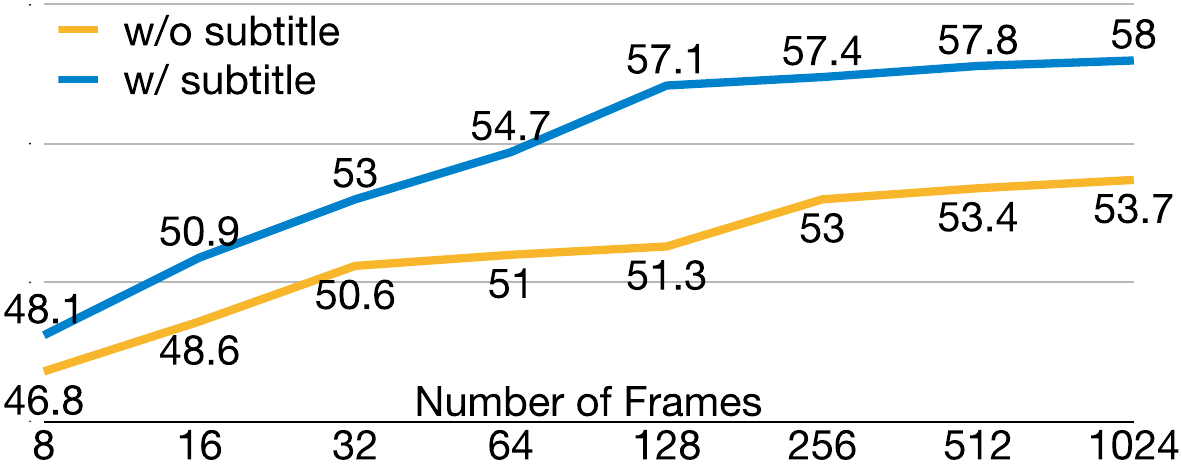}
    \caption{{Scaling video frames improves VideoMME accuracy in long category.}}
    \label{fig:scale_videomme}
  \end{minipage}%
  \hfill
  \begin{minipage}[b]{0.52\textwidth}
    \centering
\resizebox{0.96\linewidth}{!}{
\begin{tabular}{l|cccc}
\hline
\multirow{2}{*}{\begin{tabular}[c]{@{}l@{}}Training\\ Stages\end{tabular}} & \multicolumn{4}{c}{VideoMME}    \\
                                                                             & Average & Short & Medium & Long \\ \hline
1-2-3-4-5                                                                    & 57.5    & 69.3  & 56.1   & 47.0 \\
1-2-4-(3\&5)                                                                 & 55.9    & 67.4  & 54.1   & 46.1 \\
4-1-2-3-5                                                                    & 56.0    & 69.2  & 54.1   & 44.5 \\
4-1-2-(3\&5)                                                                 & 55.3    & 67.2  & 53.6   & 45.1 \\ \hline
\end{tabular}}
    \captionof{table}{{Ablations on various training stage settings on VideoMME (without subtitle). (3\&5) means joint training of the stage 3 and 5.}}
    \label{tab:ablation-trainingstage}
  \end{minipage}
\end{figure}

In this work, we introduce LongVILA, a comprehensive solution for long-context VLMs. For training {\bf pipeline}, we implement a five-stage training curriculum as Figure~\ref{fig:training-pipeline}: (1) multi-modal alignment, (2) large-scale pre-training, (3) short supervised fine-tuning, (4) context extension for LLMs, and (5) long supervised fine-tuning. For {\bf system}, we establish an efficient and user-friendly framework, namely Multi-Modal Sequence Parallelism (MM-SP), which supports training and inferencing memory-intensive long-context VLMs.

{LongVILA-7B presents strong performance on 9 popular benchmarks, \eg 65.1\% on VideoMME~\citep{video-mme} with subtitle. The LongVILA model, trained on 2048 frames, achieves 99.8\% accuracy in the needle-in-a-haystack experiments with 6,000 frames, with a context length of more than 1 million tokens.} In ablations, by increasing the number of video frames using LongVILA, the performance on VideoMME in long videos consistently improves (Figure~\ref{fig:scale_videomme}). Our MM-SP system can efficiently scale the context length up to 2 million tokens without gradient checkpointing, achieving 2.1$\times$ to 5.7$\times$ speedup compared to ring style sequence parallelism, and 1.1$\times$ to 1.4$\times$ compared to Megatron with a hybrid context parallelism and tensor parallelism.

\section{Related works}
\label{sec:related_work}

\paragraph{Visual language model architecture.}

There are two predominant designs for VLMs: the encoder-decoder architecture (\eg, LLaVA~\citep{liu2023llava}, PaLM-E~\citep{driess2023palm}) and the decoder-only architecture (\eg, Fuyu~\citep{fuyu-8b}, Chameleon~\citep{team2024chameleon}). Encoder-Decoder VLMs connect the vision encoder to the LLM decoder through a multi-modal projector. Certain multi-modal projectors, such as spatial pooling and Q-former, significantly reduce the number of tokens per image or video frame, thereby lowering the computational burden on the LLM decoder. In contrast, decoder-only LLMs typically process raw patches as input without hierarchical token pooling, making it more challenging to reduce the token count for each image or frame. In this work, we build on VILA~\citep{lin2023vila} as our foundation. It is worth noting that enhanced variants of VILA exist, such as VILA$^2$\citep{fang2024vila} for improved performance and X-VILA\citep{x-vila} for cross-modality understanding, reasoning, and generation. For our model architecture and training pipeline, we adhere to the standard VILA-1.5 version.

\paragraph{Sequence parallelism and hybrid strategy.}

Long-context training examples often exceed the memory capacity of a single device. To address this issue, the sequence parallelism paradigm has been widely adopted in the text-only LLM community, distributing a single sequence across multiple devices. {Specifically, Ring-style systems~\citet{li2021sequence, li2023lightseq, liu2023ring} use Point-to-Point (P2P) communication primitives to collectively compute the attention module,
while DeepSpeed-Ulysses~\citet{jacobs2023deepspeed} employs an All-to-All (A2A) primitive to alternate between sharding the sequence dimension and the attention head dimension during attention computation. Ulysses generally achieves higher throughput than Ring-style SP due to its more efficient A2A communication primitive and larger, unsegmented computation blocks. However, its scalability is limited by the number of attention heads.
Recently, USP \citep{USP} was introduced as the first to integrate Ring-style SP and Ulysses SP, combining the strengths of both approaches. LoongTrain \citep{LoongTrain} further optimizes communication and placement strategies to enhance training efficiency.
Following \citep{USP, LoongTrain}, we extend the system to multi-modal scenarios to accommodate complex attention masks and variable-length input sequences. Our work is the first to design and implement a sequence parallelism system for visual language models.
}

\section{LongVILA Training Pipeline}


{As shown in Figure~\ref{fig:training-pipeline},
in our pipeline, there are five training stages, {\em i.e.}, Stage 1: multi-modal alignment, Stage 2: large-scale pre-training, Stage 3: supervised fine-tuning, Stage 4: context extension for LLM, Stage 5: long supervised fine-tuning. Stage 1, 2, and 3 follow VILA~\citep{lin2023vila}, to firstly bridge the gap between LLM and vision encoder, and then pre-training on larger datasets. In Stage 1, only the multi-modal projector is trainable with others frozen. In Stage 2, we freeze the vision encoder and training LLM and the multi-modal projector. In Stage 3, we fully fine-tune the model for short data instruction following, {\em e.g.}, image and short video datasets.
Afterwards, we extend the context length of LLM with text-only dataset in a continued pre-training manner in Stage 4. In Stage 5, {we adopt our MM-SP system (\S\ref{sec_system}) to }enhance the instruction following abilities by long video supervised fine-tuning. It is noted that all parameters are trainable in the final stage.}

\subsection{Stage1\&2\&3: Alignment, Pre-training, and Short Supervised Fine-tuning}
We first use open-sourced image and video caption datasets to train the multi-modal projector in stage (1) to conduct the multi-modal alignment. Note that, following~\citep{lin2023vila}, both vision encoder and LLM decoder are frozen at this stage. After that, we conduct large-scale pre-training to learn general multi-modal capacity at scale. To improve the quality of large open-sourced datasets, we follow VILA$^2$~\citep{fang2024vila} to relabel COYO-25M~\citep{lin2023vila,kakaobrain2022coyo-700m} with VILA-1.5-40B~\citep{lin2023vila}. 
The supervised fine-tuning process incorporates mixed data types, including both images and videos. For short video comprehension, we utilize open-source video instruction-following datasets, \eg YouCook2~\citet{youcook2} and ShareGPTVideo~\citet{zhang2024direct}. In experiments, our model is based on Qwen2-1.5B and Qwen2-7B~\citep{qwen2}.

\begin{figure}[t]
    \centering
    \includegraphics[width=\linewidth]{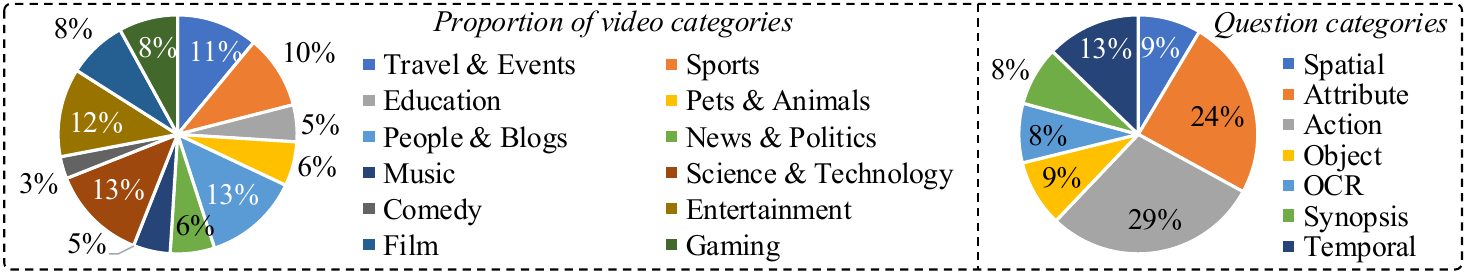}
    \caption{{The proportion of question and video categories in our LongVILA$\_$sft dataset. We have 15,292 videos in total. For each video, there are one sample for captioning and the other question.}}
    \label{fig:longvila-sft-data}
\end{figure}
\begin{figure}[t]
    \centering
    \includegraphics[width=\linewidth]{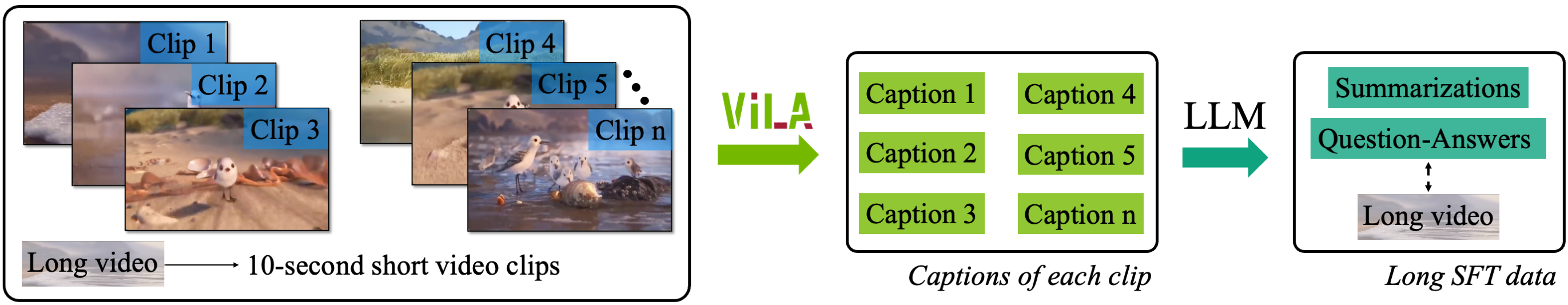}
    \caption{{The pipeline for generating instruction-following data from long videos. The process begins by segmenting the long video into short clips, each approximately 10 seconds in length. These clips are individually annotated with captions using the VILA-1.5 model. Subsequently, an LLM is employed to generate question-and-answer pairs based on the captions of these clips. Generated questions include summarization and other inquiries pertinent to the content of long videos.
}}
    \label{fig:longvideo-data}
\end{figure}

\subsection{{Stage4: Context Extension for LLMs}}
{
Our empirical research indicates that extending the context length of LLMs is essential prior to engaging in supervised fine-tuning with long video datasets. Following Stage 2 of our methodology, we execute a continuation of pre-training on the LLM to enhance its context length to 262,144, utilizing a total of 17B tokens. We employ a progressive training schedule, incrementally increasing the context length from 8,192 to 65,536, and ultimately to 262,144, utilizing the SlimPajama dataset~\citep{cerebras2023slimpajama} in accordance with the methodology outlined by \citep{long-context-data-engineering}.

Furthermore, we augment the base frequency of the Rotary Position Embeddings (RoPE) as described in \citep{rope} during the fine-tuning phase. Sequence parallelism is implemented for the training at the 262,144 context length. We use low-rank adaptation for context extension fine-tuning~\citep{longlora}. These processes collectively require approximately 336 GPU hours on machines equipped with 80GB A100 GPUs.
}

\subsection{{Stage5: Long Supervised Fine-tuning}}
\label{sec:method-stage4}

\textbf{{Long video instruction following}}
To facilitate the fine-tuning of long videos, we constructed a new, dedicated dataset for long video training, each consisting of 15,292 videos. We use the original long videos from the Shot2Story dataset~\citep{han2023shot2story20k}. Each video includes different questions and answers: one for generating captions and another for answering questions, enabling diverse applications in video understanding. Figure~\ref{fig:longvideo-data} illustrates the process for generating instruction-following datasets from long videos. Initially, the long video is segmented into shorter clips, each approximately 10 seconds in duration. These clips are then independently annotated with descriptive captions utilizing the VILA-1.5 model. Subsequently, an LLM is employed to generate question-and-answer pairs derived from the captions of these clips. The generated questions encompass summarization and other queries relevant to the comprehensive understanding of long video content. 

As in Figure~\ref{fig:longvila-sft-data}, the left chart categorizes videos into several domains, including Travel \& Events, Sports, Education, Pets \& Animals, People \& Blogs, News \& Politics, Music, Science \& Technology, Comedy, Entertainment, Film, and Gaming, ensuring a wide-ranging representation of video content. The right chart breaks down the categories of questions into Spatial, Attribute, Action, Object, OCR, Synopsis, and Temporal, reflecting the variety of inquiries and cognitive tasks that the dataset can address. This dataset provides a rich resource for advancing the understanding and processing of long video formats in supervised fine-tuning.

{Once we acquired the long video dataset, applying it for supervised fine-tuning introduced new challenges, primarily due to the substantial number of frames in each sample—often ranging in the hundreds or even thousands. For instance, a single sequence from 1400 video frames can encompass around 274k tokens. Existing data-parallel training systems struggle to handle such extensive contexts. We developed the MM-SP system (Section \ref{sec_system}) to efficiently train long-context VLMs.}

\section{Multi-Modal Sequence Parallelism}
\label{sec_system}

Training long-context Vision-Language Models (VLMs) results in substantial memory demands. The most widely used open-source solution, fully sharded data parallelism, does not distribute the activations generated by a single sequence, making it unsuitable for our needs. Consequently, we developed a custom system based on sequence parallelism~\citep{li2021sequence, li2023lightseq, liu2023ring, jacobs2023deepspeed}, a technique commonly employed in existing foundation model systems to optimize text-only LLM training. However, we discovered that existing systems are neither efficient nor scalable enough to handle our long-context VLM workloads.

\begin{figure}
    \centering
    \includegraphics[width=\linewidth]{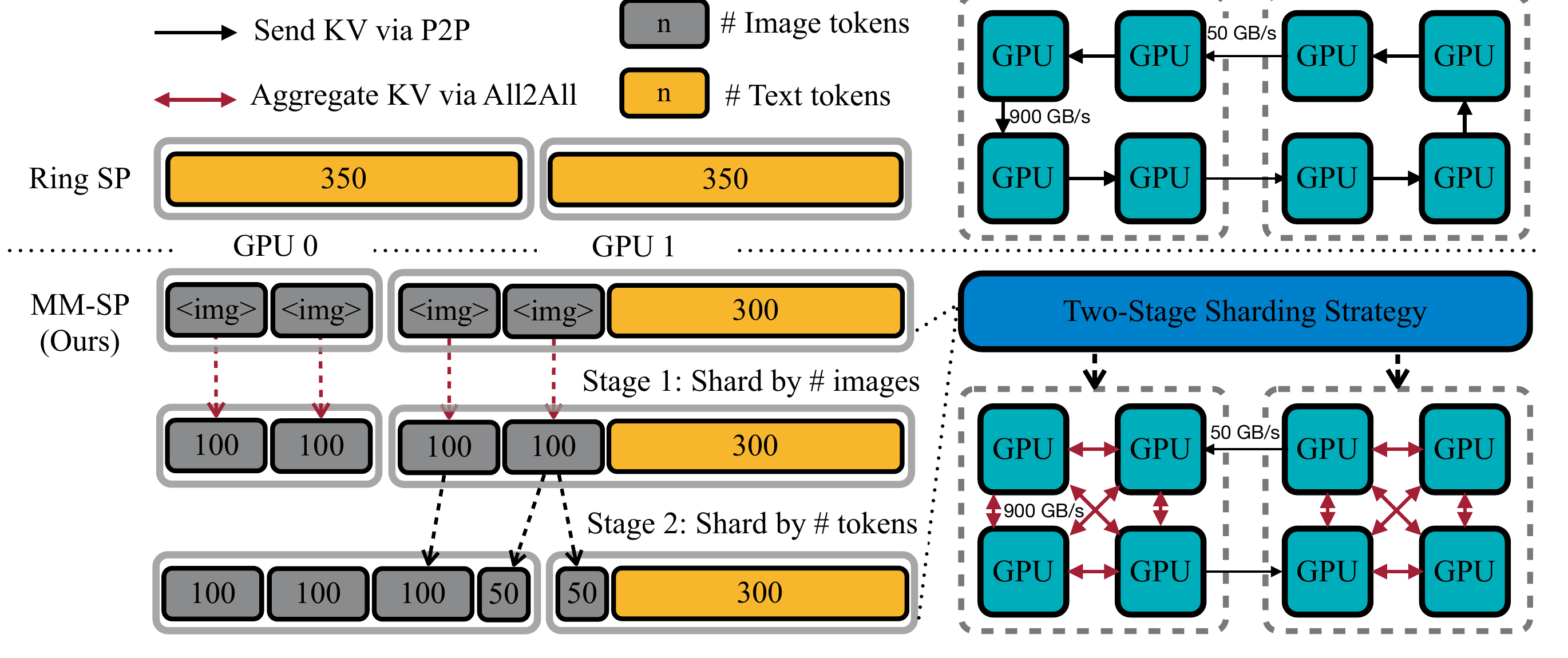}
\caption{{Sharding strategy and communication pattern of MM-SP. For sharding strategy, Ring SP is designed for text-only modalities, without optimization for the workload of an image encoder. Our MM-SP implements a novel sharding strategy that balances the computational load between the image encoder and the language modeling stages. 
For communication pattern, Ring SP~\citep{liu2023ring, li2023lightseq} (top) relies on P2P communication for both intra-node and inter-node settings, resulting in underutilization of intra-node bandwidth.  MM-SP  (bottom) adopts 2D-Attention \citep{USP, LoongTrain} mechanism which utilizes intra-node All-to-All (All2All) and inter-node Point-to-Point (P2P) communication to transfer keys and values (KV), enhancing the efficiency of intra-node NVLink utilization. The bandwidth is for H100. }}
    \label{fig:system_overview}
\end{figure}

\subsection{Limitations of Existing Systems}
\label{sec:system_limitation}

\textbf{Modality heterogeneity.} In text-only LLMs, sequences are processed by a single tokenizer into tokens, allowing for straightforward distribution of tokens across multiple GPUs. However, VLMs incorporate an encoder architecture where non-text data is initially represented by a placeholder token (\eg $<$img$>$) and subsequently encoded into multiple real tokens during training. For instance, a single video frame typically requires around 256 tokens~\citep{lin2023vila}. Due to the differing processing requirements of visual and text modalities, a simplistic implementation that treats placeholder tokens the same as text tokens leads to an imbalance in GPU workloads (Figure~\ref{fig:system_overview}).

\textbf{Networking heterogeneity.}
Our multi-modality comprises extremely long videos (Figure~\ref{fig:training-pipeline}), which requires employing sequence parallelism in a \textit{multi-node} setting. In a multi-node setting, inter-node and intra-node network bandwidth differs significantly. For example, the NVIDIA DGX H100 utilizes NVLink at 900 GB/s for intra-node GPU communication and InfiniBand at 50 GB/s for inter-node GPU communication (single path), resulting in an 18$\times$ difference in bandwidth. Previous work, Ring-Style sequence parallelism~\citep{li2021sequence, li2023lightseq, liu2023ring, ring-flash-attention} ignores the heterogeneous networking feature on GPUs and utilizes P2P communication in both inter-node and intra-node settings. This design induces excessive communication costs where they usually attempt to overlap them into computation. However, we found that this design cannot always hide the overhead, and even slows down the computation kernel (Table~\ref{tab:ring_style_computation}).

\textbf{Limited maximal sequence length.}
DeepSpeed-Ulysses~\citep{jacobs2023deepspeed} presents a potential solution to the communication challenges in ring-style sequence parallelism by employing All-to-All communication primitives, which reduce the overall communication volume. However, this approach has its limitations. The design relies on parallelizing along the attention head dimension rather than the sequence dimension during attention computation. As a result, DeepSpeed-Ulysses cannot scale effectively beyond the number of attention heads. For instance, the Llama-3 8B model uses Grouped Query Attention (GQA) with 8 Key-Value heads, which restricts the maximum sequence parallelism degree to 8. Even when using replication for Key-Value heads, which introduces additional communication overhead~\citep{li2023lightseq}, the highest achievable sequence parallelism degree is still limited to 32 (the number of Query heads). This constraint is insufficient for handling extremely long sequences, such as full-length movies.

\subsection{Multi-Modal Sequence Parallelism Training Mode}
\label{sec:method_training}
After identifying the limitations in existing systems, we conclude that an ideal multi-modal sequence parallelism approach should prioritize efficiency and scalability by addressing both modality and network heterogeneity, and should also be capable of scaling beyond the number of attention heads. 
{To achieve this, we adopt 2D-attention  \citep{USP, LoongTrain} mechanism for sequence parallelism. For instance, as illustrated on the left in Figure \ref{fig:system_overview}, to enable an 8-degree sequence parallelism across 2 nodes, we construct a 4$\times$2 communication mesh using 2D-SP. In this setup, the A2A process group, with a size of 4, distributes the QKV tensors according to the head dimension and re-partitions them along the sequence dimension within each node. Simultaneously, the P2P process group, with a size of 2, transfers the partitioned KV chunks between nodes. 
 Additionally, to further explain how the 2D-attention mechanism operates, we depict the attention computation schedule using different methods in Figure \ref{fig:system_2d_attn}.}

\begin{figure}
    \centering
    \includegraphics[width=0.95\linewidth]{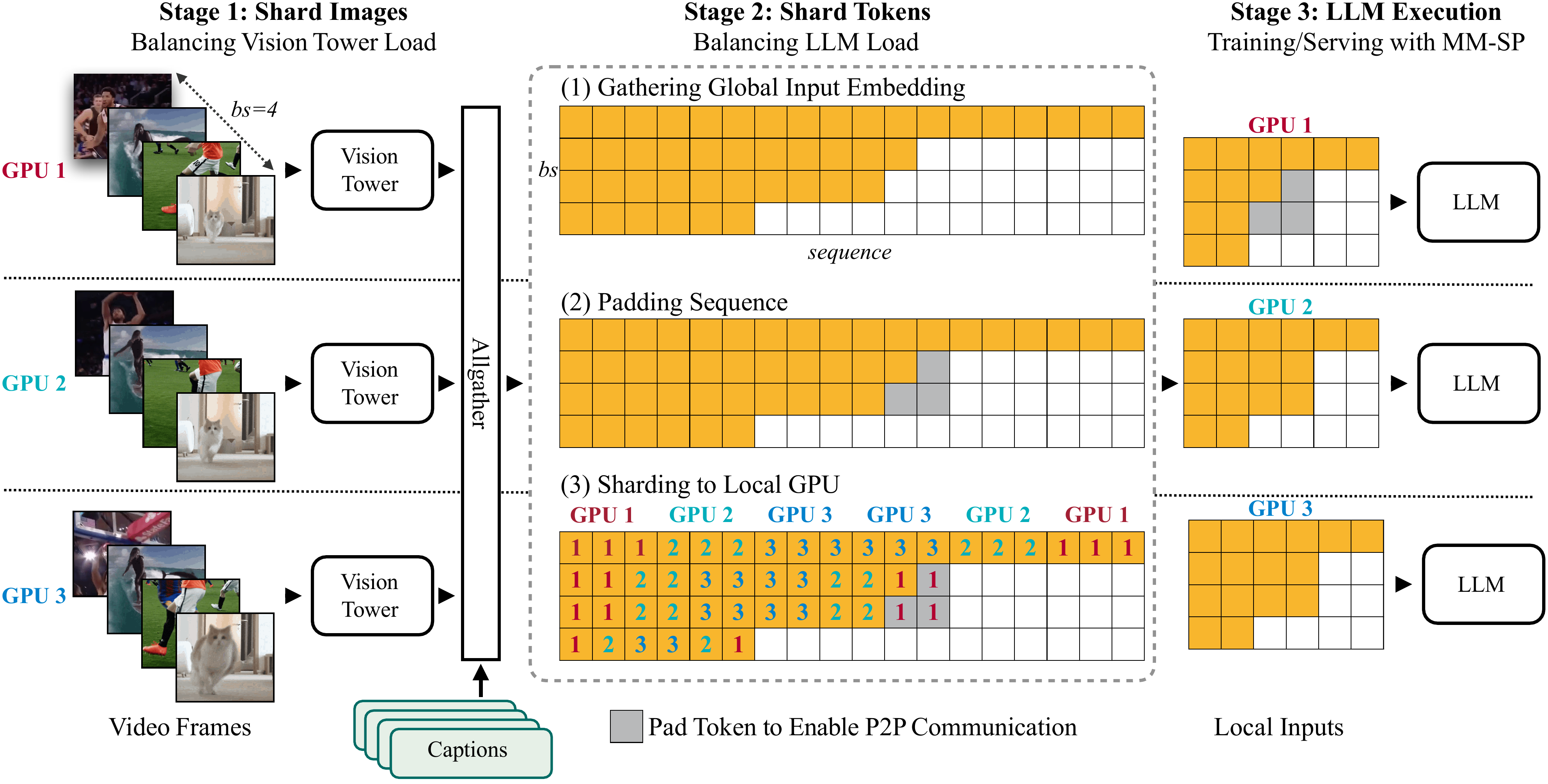}
    \caption{Workflow of Multi-Modal Sequence Parallelism with a Batch Size (bs) of 4 and a Sequence Parallel Size (SP\_Size) of 3. To accomodate multi-modal inputs, we developed a customized sharding strategy that ensures balanced workload distribution and compatibility with {SP communucation}.}
    \label{fig:system_workflow}
\end{figure}

\textbf{MM-SP workflow.}
To address the challenge of modality heterogeneity, we propose a two-stage sharding strategy that optimizes the compute workload for both image encoding and language modeling stages. As illustrated in Figure \ref{fig:system_workflow}, the process begins by evenly distributing images (e.g., video frames) across devices within the sequence parallelism (SP) process group, thereby achieving load balancing during the image encoding stage. In the second stage, we aggregate global vision and text inputs for token-level sharding. To support ring-based attention, sequences are extended with arbitrary dummy tokens, ensuring that each sequence can be evenly divided according to the ring-based SP degree. This adjustment maintains consistency with the original approach by modifying label inputs to ignore padded tokens during loss calculation. We implement a balanced sharding strategy that distributes the context to each rank from both ends, ensuring equal computation across ranks. The effectiveness of this strategy will be demonstrated later (Table~\ref{tab:two_stage}). Since this redistribution is performed only once during training, the overhead is minimal. Finally, the balanced local inputs are processed by the LLM backbone, utilizing 2D-Attention to achieve efficient sequence parallelism.

\begin{table}[t]
\centering
\caption{\dacheng{The forward and backward attention kernel wall-clock time with or without the overlapping design (Unit: $\mu$s). The communication overlap design in Ring-style SP \textbf{slows down} the attention kernel by occupying streaming multiprocessor (SM) resources.}}
\small
\resizebox{0.9\columnwidth}{!}{
\color{DLTODO}{
\begin{tabular}{llllll}
\toprule
Seq. length  & 4K &     8K & 16 K & 24K & 32K\\
\midrule
forw. w/o & 29.5 & 49.3& 122.1 & 239.2 & 402.9 \\
forw. w/  & 35.0 (+18.6\%) & 54.6 (+10.7\%) & 131.2 (+7.5\%) & 250.9 (+4.8\%) & 420.1 (+4.2\%) \\
backw. w/o & 77.7 & 123.3 & 362.9 & 730.0 & 1218.9 \\
backw. w/ & 82.2 (+5.8\%) & 129.8 (+5.3\%)& 367.0 (+1.1\%) & 743.2 (+1.8\%)& 1225.3 (+0.5\%)\\
\bottomrule
\end{tabular}}}
\label{tab:ring_style_computation}
\end{table}

\subsection{Multi-Modal Sequence Parallelism Inference Mode}

The model we developed through sequence parallelism training is capable of handling long-context multi-modal downstream tasks. However, the most commonly used inference system, built on HuggingFace Transformers, typically operates on a single GPU. This lack of distributed implementation limits the maximum sequence length that can be processed during inference. The most straightforward solution within HuggingFace Transformers is to use its pipeline parallelism inference feature, which shards a single model across multiple devices on a layer-by-layer basis~\citep{huang2019gpipe, narayanan2019pipedream}. However, this approach is inefficient, as it only activates one GPU at a time. Additionally, it struggles to support long sequence because the first device must store large input embeddings and images, creating a memory bottleneck.
\begin{wrapfigure}{r}{0.4\textwidth}
  \centering
  \includegraphics[width=0.4\textwidth]{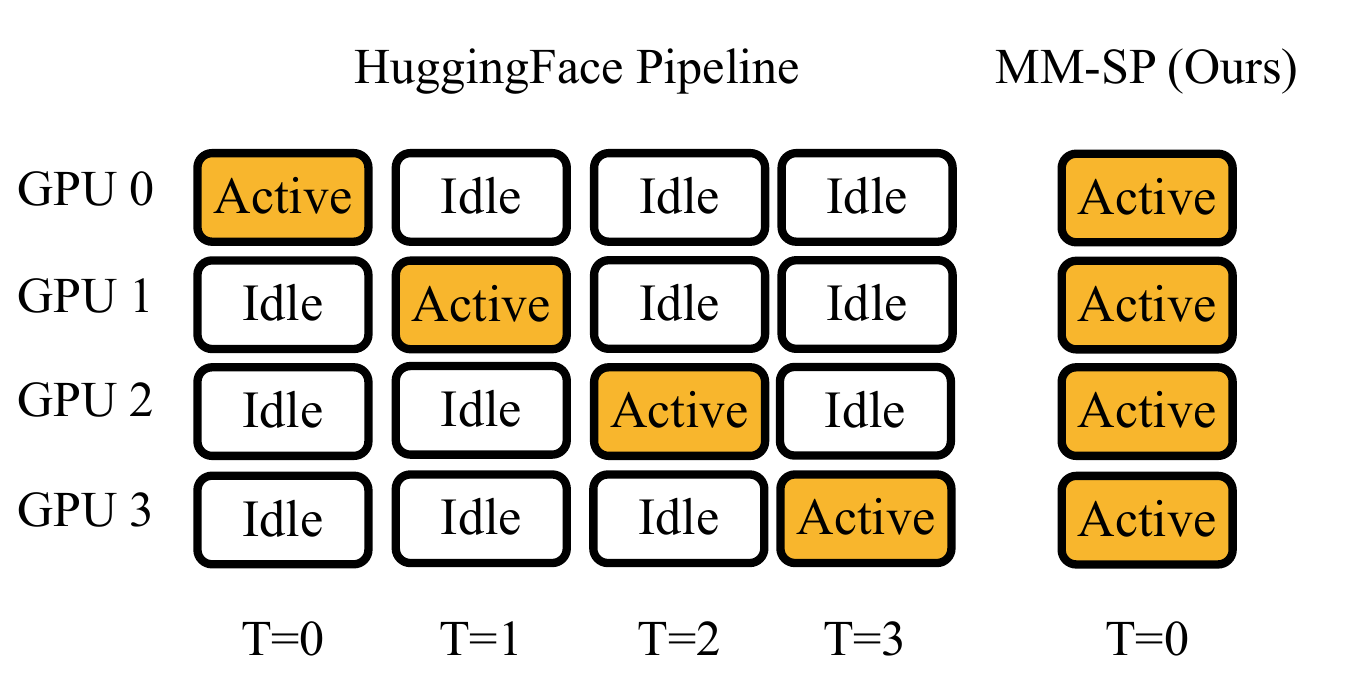}
  \caption{Inference scheduling comparison between HuggingFace Pipeline and MM-SP, illustrated with 4 GPUs. MM-SP utilizes all GPUs concurrently.
  }
\label{fig:inference_vs_hf}
\end{wrapfigure}

To address these limitations, we implemented sequence parallelism for distributed inference in VLMs. Unlike the training mode, the inference system additionally manages tensors, such as input tokens and position encodings, that progressively change during the decoding phase~\citep{yu2022orca}. It detects signals from the machine with the last token to terminate the distributed process appropriately. Compared to HuggingFace's pipeline parallelism strategy, our inference mode is more efficient, as all devices participate in computation simultaneously, accelerating the process by a factor proportional to the number of machines (Figure~\ref{fig:inference_vs_hf}). Furthermore, it is scalable, with memory evenly distributed across devices, enabling longer sequences with additional machines.

\section{Experimental Results}


\subsection{Training and inference system}

Our training and inference systems can be integrated with HuggingFace Transformers through straightforward monkey patching, in line with the popular open-source approach outlined in~\citep{zheng2023judging}. In this section, we present a quantitative evaluation of the training system's throughput, the inference system's latency, and the maximum supported sequence length.

\begin{figure}[t]
\begin{subfigure}[t]{0.48\textwidth}
  \centering
    {\includegraphics[width=1.0\textwidth]{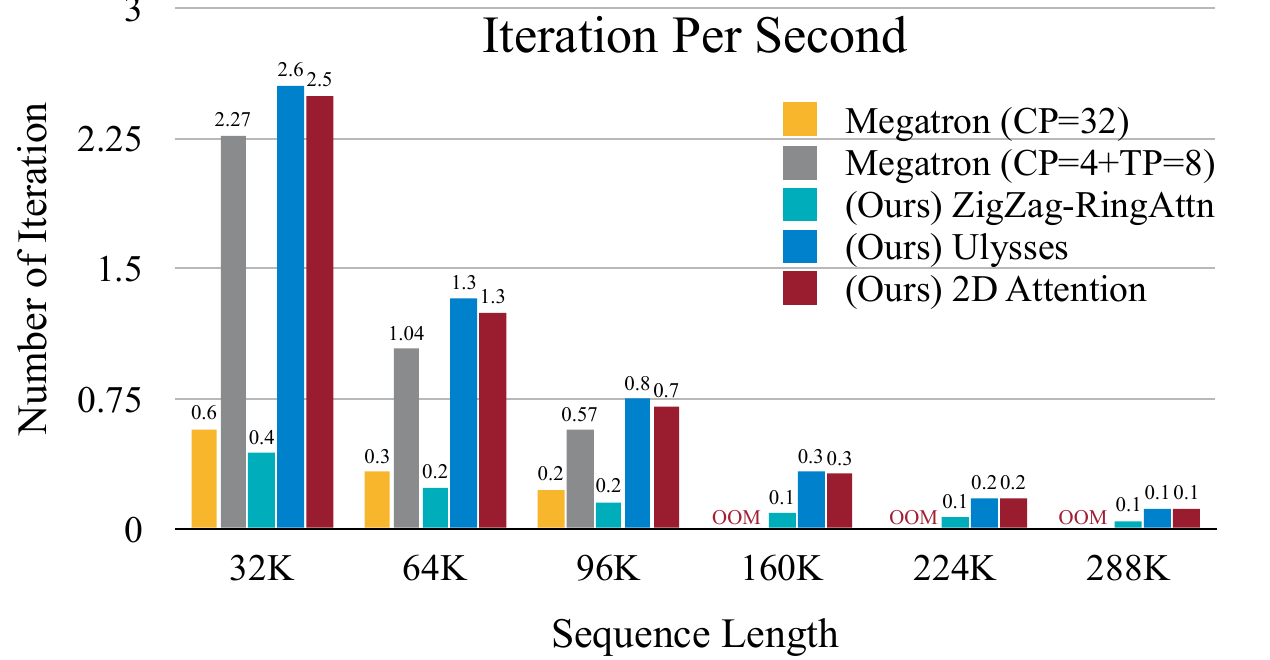}}
\end{subfigure}
\begin{subfigure}[t]{0.48\textwidth}
{\includegraphics[width=1.0\textwidth]{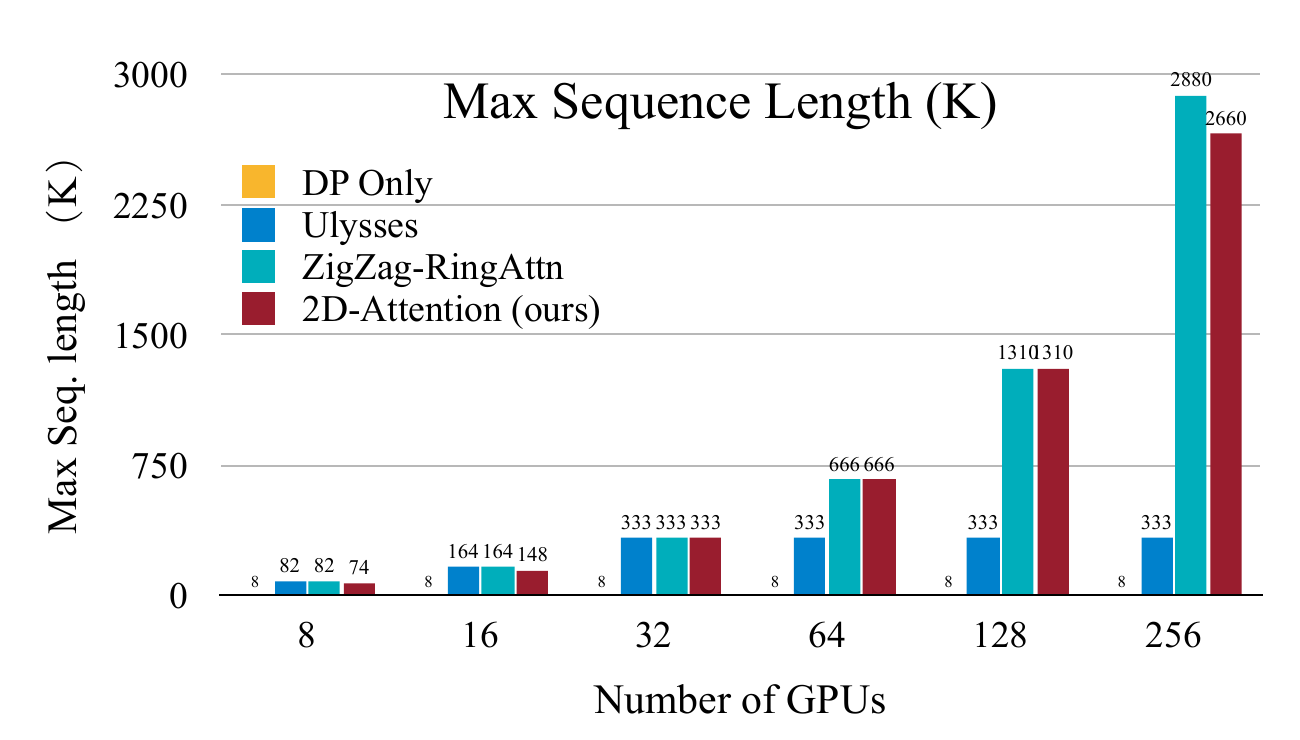}}
\end{subfigure}
\caption{Performance comparison of training systems on 32 H100 GPUs. MM-SP is as scalable as ZigZag-RingAttn, and as efficient as Ulysses.}
\label{fig:train_performance}
\end{figure}
\begin{figure}[t]
\begin{subfigure}[t]{0.48\textwidth}
  \centering
{\includegraphics[width=1.0\textwidth]{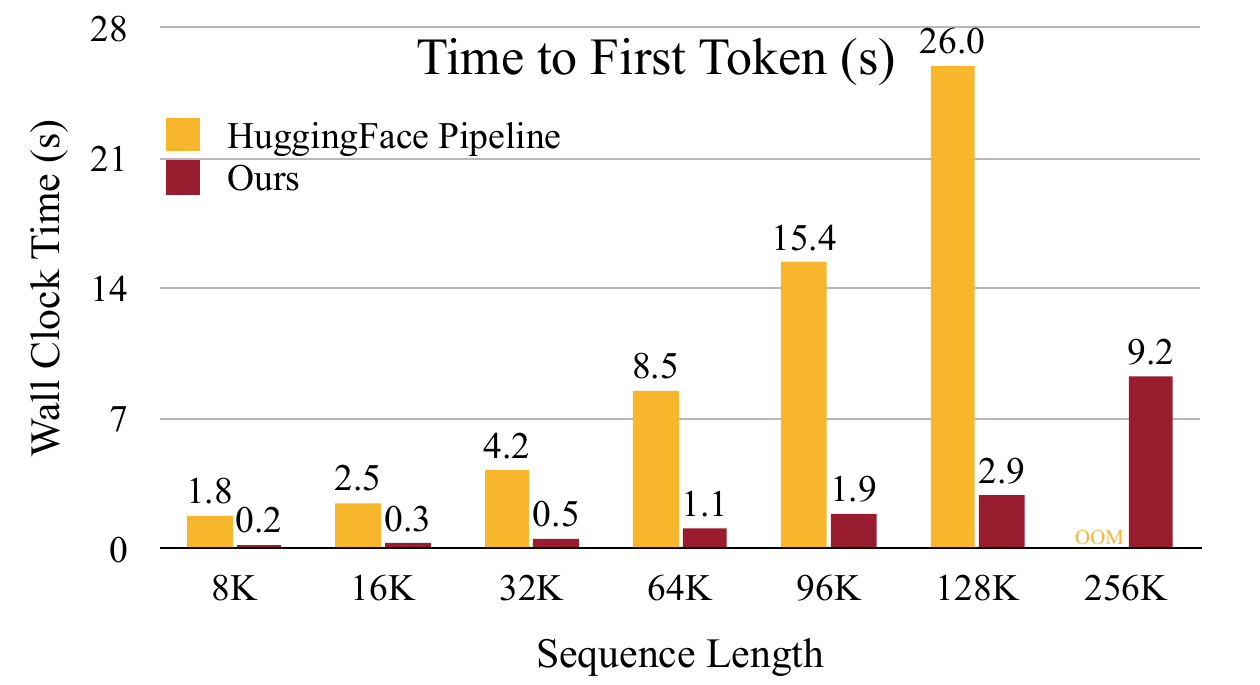}}
\end{subfigure}
\begin{subfigure}[t]{0.48\textwidth}
{\includegraphics[width=1.0\textwidth]{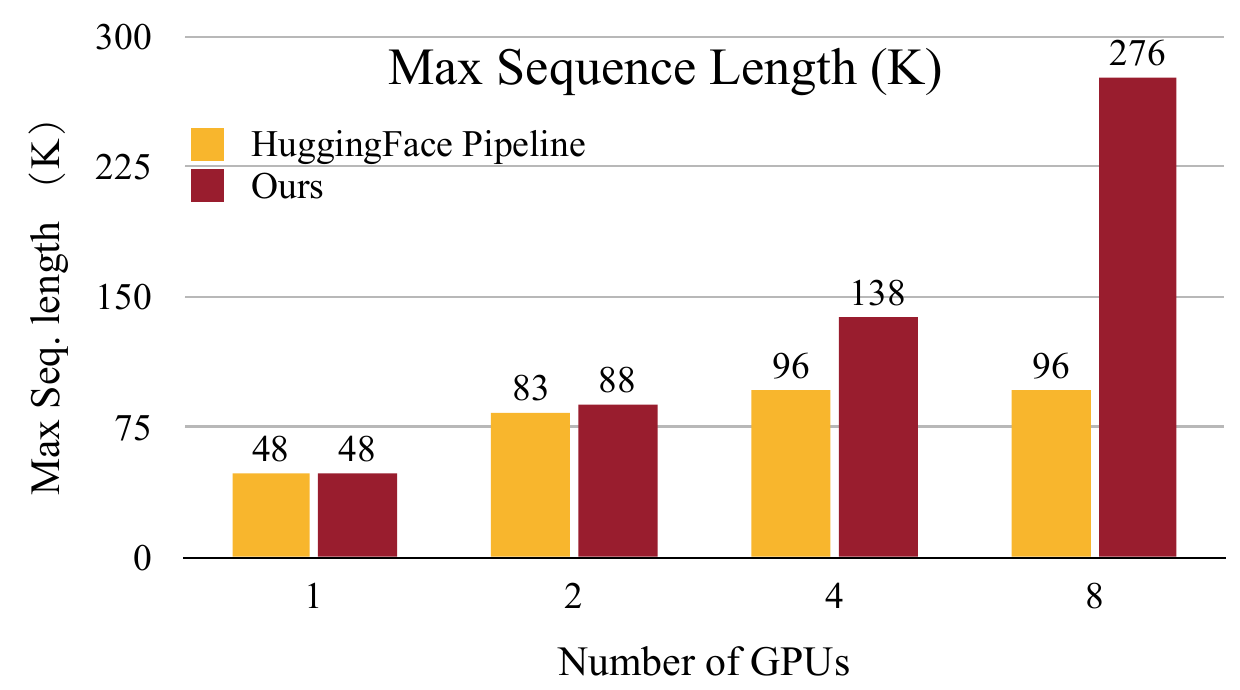}}
\end{subfigure}
\caption{Performance comparison of inference systems on 8 H100 GPUs.}
\label{fig:inference_performance}
\end{figure}

\subsubsection{Training system} 
\textbf{Baselines and hardware setup} 
For training efficiency, we compare our system with ZigZag ring-style sequence parallelism, which incorporates load balancing and GPU optimization~ (\zigzag for consistency)\citep{li2023lightseq, ring-flash-attention, liu2024world, korthikanti2023reducing}. We use a widely adopted open-source implementation~\citep{ring-flash-attention}. To reduce the memory footprint of models, gradients, and optimizer states, we employ Fully-Sharded Data Parallelism (FSDP)\citep{zhao2023pytorch} instead of Zero-3\citep{rajbhandari2020zero} (Table~\ref{tab:training_system_efficiency-appendix}). Additionally, we compare our system with the expert-designed and highly optimized Megatron-LM\citep{shoeybi2019megatron, korthikanti2023reducing} system, focusing on their implementation of sequence parallelism, termed ``context parallelism" (CP). We also evaluate a hybrid strategy that combines tensor model parallelism (TP) within a node and CP across nodes, as recommended by the Megatron-LM team for advanced usage.

We conduct most experiments on H100 nodes, each equipped with 8xH100 (80GB) GPUs interconnected via intra-node NVLink and 400 Gbps inter-node InfiniBand. For experiments involving the maximum supported sequence length during training, we extend the setup to 32 A100 nodes, each with 8xA100 (80GB) GPUs, where the conclusions are consistent with those for H100 due to the equivalent total memory. Our evaluations are based on an 8B model with a batch size of 1. Since the Megatron-LM baseline does not natively support VLM training and the visual encoder is typically orders of magnitude smaller than LLMs, we report the main results for the LLM backbone without the visual encoder. An ablation study of the visual encoder is provided in \S~\ref{sec:effect_two_stage}.

\begin{table}[t]
\begin{center}
\caption{{{Performance comparisons on 9 video benchmarks, including ActivityNet-QA~\citep{activity-qa}, EgoSchema~\citep{egoschema}, EventBench~\citep{eventbench}, LongVideoBench~\citep{longvideobench}, PerceptionTest~\citep{perpceptiontest}, MVBench~\citep{mvbench}, NExT-QA~\citep{next-qa}, VNBench~\citep{vnbench}, and VideoMME~\citep{video-mme}.}}}
\resizebox{\linewidth}{!}{
{
\begin{tabular}{l|c|cccccccccc}
\toprule
\multirow{2}{*}{Model} & \rotatebox{90}{LLM Size} & \rotatebox{90}{ActNet-QA} & \rotatebox{90}{EgoSchema}   & \rotatebox{90}{EventBench}    & \rotatebox{90}{LVideoBench} & \rotatebox{90}{PercepTest} & \rotatebox{90}{MVBench}       & \rotatebox{90}{NExT-QA}       & \rotatebox{90}{VNBench}   & \multicolumn{2}{c}{VideoMME} \\ \cline{3-12}
                       &                                                                     & test           & test        & val           & val            & val            & test          & mc            & val       & w/o sub.  & w/ sub.   \\ \midrule
GPT-4V            & -                                                                 &  57.0          &  -      &   32.6        &    61.3        &    -           &   43.5       &   -        &   -    &    59.9     &   63.3      \\
GPT-4o            & -                                                                 &  -          &   -     &     53.3      &   66.7         &     -          &   -       &    -       &    64.4   &     71.9    &   77.2      \\
Gemini-1.5-Pro            & -                                                                 &  57.5          &   72.2     &    43.2       &   64.0         &   -            &    -      &    -       &   66.7    &    75.0     &   81.3      \\ \midrule
Video-LLaVA            & 7B                                                                  & 45.3           & 38.4        & 5.9           & 37.6           & -              & 43.5          & -             & 12.4      & 39.9          & 41.6          \\
Flash-VStream          & 7B                                                                  & 51.9           & -           & -             & -              & -              & -             & 61.6          & -         & -             & -             \\
ShareGPT4Video         & 8B                                                                  & 50.8           & -           & -             & 41.8           & -              & 51.2          & -             & -         & 39.9          & 43.6          \\
VideoLLaMA2            & 7B                                                                  & 50.2           & 51.7        & 6.9           & -              & 51.4           & 54.6          & -             & 4.5       & 47.9          & 50.3          \\
VideoLLaMA2.1          & 7B                                                                  & 53.0           & 53.1        & -             & -              & 54.9           & 57.3          & -             & -         & 54.9          & 56.4          \\
Kangaroo               & 8B                                                                  & -              & 62.7        & -             & 54.8           & -              & 61.1          & -             & -         & 56.0          & 57.6          \\
PLLaVA                 & 7B                                                                  & 56.3           & -           & 28.2          & 39.2           & -              & 46.6          & -             & -         & -             & -             \\
LLaVA-OV               & 7B                                                                  & 56.7           & 60.1        & -             & 56.4           & 57.1           & 56.7          & 79.4          & 51.8      & 58.2          & 61.5          \\ \midrule
LongVILA               & 7B                                                                  & \textbf{59.5}  & \textbf{67.7} & \textbf{58.0} & \textbf{57.1}  & \textbf{58.1}  & \textbf{67.1} & \textbf{80.7} & \textbf{63.0} & \textbf{60.1} & \textbf{65.1} \\ \bottomrule
\end{tabular}}
}
\label{tab:9-benchmarks}
\end{center}
\end{table}

\textbf{Throughput} 
Figure \ref{fig:train_performance} presents throughput results measured as iteration per second over 32 H100 GPUs. These results were obtained after 10 warmup iterations and averaged over 5 iterations to minimize variance. Our system achieves a speedup of $2.1\times$ to $5.7\times$ compared to \zigzag, and performs on par with DeepSpeed-Ulysses. When compared to the more optimized ring-style sequence parallelism in Megatron-LM CP, our method shows a $3.1\times$ to $4.3\times$ speedup. This highlights that our {system design} effectively addresses the issues inherent in ring-style sequence parallelism, as in $\S~\ref{sec:method_training}$. Furthermore, our system achieves a $1.1\times$ to $1.4\times$ speedup compared to Megatron-LM’s hybrid strategy. Note that our system is currently implemented in Triton~\citep{tillet2019triton}, and further porting it to C++ could yield even greater speedup. Additionally, we observed that the Megatron-LM system supports a significantly lower maximum sequence length, which is why its results are not in the next section. We observe similar observations using 8 H100 nodes (Table~\ref{tab:training_system_efficiency_64GPUs}).

We evaluate the maximum sequence length supported by a fixed number of GPUs by progressively increasing the per-GPU sequence length from 1k to 10k until an out-of-memory error occurs. The results are summarized in Figure~\ref{fig:train_performance}. To ensure a fair comparison, activation checkpointing is disabled. Vanilla data parallelism fails to scale for long videos at larger cluster sizes. DeepSpeed-Ulysses partitions based on attention heads, which limits its ability to scale to higher context lengths, as the 8B model has only 32 attention heads. Consequently, our approach supports approximately $8\times$ higher context lengths when scaled to 256 GPUs. Additionally, our system achieves a similar context length scaling as \zigzag, with support for over \textbf{2 million} context length on 256 GPUs.

In summary, our training system combines the best of both worlds—it achieves scalability comparable to \zigzag while maintaining the throughput of DeepSpeed-Ulysses. Additionally, it delivers a 1.3$\times$ speedup and supports 2.5$\times$ longer context lengths compared to the highly optimized Megatron-LM.

\subsubsection{Inference system} 
We evaluated our inference system against HuggingFace Pipeline parallelism using a single node with 8 H100 GPUs and the 8B model (Figure \ref{fig:inference_performance}). Our system achieves an $8.2\times$ speedup compared to HuggingFace Pipeline on 8xH100 GPUs. This significant improvement is primarily due to HuggingFace Pipeline inference activating only one GPU at a time, whereas our method leverages all GPUs to compute jointly. Figure~\ref{fig:inference_performance} compares the maximum supported sequence length, where our method supports sequences that are $2.9\times$ longer than those supported by HuggingFace Pipeline. Specifically, during 96K sequence length inference, 
HuggingFace Pipeline stores 80GB of activations on the first GPU and only 18GB on the remaining GPUs. This imbalanced allocation of activations limits the maximum supported sequence length.

\subsubsection{Effect of two-stage sharding}
\label{sec:effect_two_stage}

We evaluate the impact of our two-stage sharding strategy using a video captioning dataset~\citep{chen2024sharegpt4video}. We compare our two-stage sharding to a one-stage baseline that only distributes workload based on the number of images. We measure the time per iteration across different numbers of H100 GPUs. For $k$ GPUs, we use $k$ images per video and a batch size of $k$. The results, shown in Table~\ref{tab:two_stage}, indicate a speedup ranging from 1\% to 7\%. This improvement is primarily observed in longer captioning tasks, where the baseline suffers from workload imbalance due to the lack of sharding based on the number of text tokens.

\begin{table}[t]
\begin{center}
\caption{{Performance comparison on VideoMME~\citep{video-mme} benchmark in details.}}
\resizebox{\linewidth}{!}{
\begin{tabular}{l|c|c|cccc|cccc}
\toprule
\multirow{2}{*}{Model} & \multirow{2}{*}{\begin{tabular}[c]{@{}c@{}}LLM\\ Size\end{tabular}} & \multirow{2}{*}{Frames} & \multicolumn{4}{c|}{\textit{w/o subtitle}} & \multicolumn{4}{c}{\textit{w subtitle}} \\
                       &                       &                         & Overall    & Short    & Medium    & Long   & Overall   & Short   & Medium   & Long   \\ \midrule
Video-LLaVA            & 7B                    & 8                       & 39.9       & 45.3     & 38.0      & 36.2   & 41.6      & 46.1    & 40.7     & 38.1   \\
SliME                  & 8B                    & 8                       & 45.3       & 53.3     & \underline{55.4}      & 39.8   & 47.2      & 55.4    & 44.4     & 41.7   \\
ShareGPT4Video         & 8B                    & 16                      & 39.9       & 48.3     & 36.3      & 35.0   & 43.6      & 53.6    & 39.3     & 37.9   \\
VideoChat2             & 7B                    & 16                      & 39.5       & 48.3     & 37.0      & 33.2   & 43.8      & 52.8    & 39.4     & 39.2   \\
VideoLLaMA2            & 7B                    & 16                      & 47.9       & 56.0     & 45.4      & 42.1   & 50.3      & 59.4    & 47.6     & 43.8   \\
Chat-Univi-v1.5        & 7B                    & 64                      & 40.6       & 45.7     & 40.3      & 35.8   & 45.9      & 51.2    & 44.6     & 41.8   \\
Kangaroo               & 8B                    & 64                      & \underline{56.0}       & 66.1     & 55.3      & 46.7   & \underline{57.6}      & {68.0}    & 55.4     & {49.3}   \\ 
ShareGemini            & 7B                    & 64                      & 43.2       & 49.1     & 41.3      & 39.1   & 47.9      & 49.1    & 47.3     & 43.4   \\
LongVA                 & 7B                    & 128                     & 52.6       & 61.1     & 50.4      & 46.2   & 54.3      & 61.1    & 53.6     & 47.6   \\
\midrule
InternVL-V1.5          & 20B                   & 10                      & 50.7       & 60.2     & 46.4      & 45.6   & 52.4      & 61.7    & 49.1     & 46.6   \\
VITA                   & 8x7B                  & 20                      & 55.0       & 64.2     & 53.3      & \underline{47.6}   & 57.6      & 67.9    & 55.3     & \underline{49.6}   \\ 
Video-CCAM             & 14B                   & 96                      & 53.9       & 62.1     & 52.8      & 47.0   & 56.1      & 63.9    & \underline{55.9}     & 48.3   \\
\midrule

\multirow{2}{*}{{LongVILA}}              & {1.5B}                    & \multirow{2}{*}{{256}}                     &  {53.6}      &   {\underline{66.2}}   &   {49.3}    &  {45.3}  &   {57.5}    &  {\underline{70.2}}   &   {54.1}   &  {48.2}  \\ 
              & {7B}                    &                      & {\textbf{60.1}}       & {\textbf{69.0}}     & {\textbf{58.3}}      & {\textbf{53.0}}   & {\textbf{65.1}}      & {\textbf{72.9}}    & {\textbf{64.9}}     & {\textbf{57.4}}   \\ \bottomrule

\end{tabular}}
\label{tab:video-mme}
\end{center}
\end{table}

\subsection{General Video Understanding}

Table~\ref{tab:9-benchmarks} presents the performance of LongVILA, comparing to state-of-the-art models~\citep{video-llava, slime, chen2024sharegpt4video, videochat, videollama2, chat-univi, longva, sharegemini, intervl, vita, video-ccam, kangaroo, pllava, llava-onevision, flash-vstream} on popular video benchmarks. LongVILA-7B achieves strong performance across all these benchmarks.
Table~\ref{tab:video-mme} compares their effectiveness across short, medium, and long video lengths, as well as overall performance on VideoMME. {LongVILA, utilizing 256 frames, achieves an overall score of 60.1 / 65.1 without / with subtitle, which are competitive results. We include LongVILA-1.5B (starting from Qwen2-1.5B~\citep{qwen2}) for evaluation, which is also competitive. We provide a detailed model complexity of LongVILA among various model size, number of frames, context length, latency and FLOPs in Table~\ref{tab:latency_flops_profile} in the appendix.
We also do the ablation on training schedules in Table~\ref{tab:ablation-trainingstage}.}

\subsection{{Needle-in-a-Haystack}}
{In Figure~\ref{fig:needle-in-a-haystack}, we present the results of the Needle in a Haystack experiment for long videos. Following the methodology established in the existing literature~\citep{longva}, we prepared a long video and sampled a fixed number of frames. We inserted specifically designed images at various depths and tasked the model with answering corresponding questions. The 32-frame baseline model (left) was unable to accurately retrieve the correct images beyond 32 frames. {In contrast, the LongVILA model (right) demonstrated 99.8\% accuracy across 6,000 frames, which contains more than 1 million tokens. To our best knowledge, this is the first VLM which can handle 1 million context length.}}

\section{Conclusion}

We introduce LongVILA, a comprehensive full-stack solution for long-context visual language models, encompassing model training pipeline and distributed system. Based on our curated long video datasets and five-stage training pipeline, {our LongVILA model extends the feasible frame count from 8 to 2048, precisely capturing fine-grained information from 2-hour needle-in-a-haystack videos, with more than 1 million tokens Our LongVILA-7B model achieves strong performance across popular video benchmarks, especially on VideoMME, \eg 65.1\% accuracy with subtitle.} Our system efficiently scales context length up to 2 million tokens, achieving speedups of 2.1$\times$ to 5.7$\times$ compared to ring sequence parallelism and 1.1$\times$ to 1.4$\times$ compared to a hybrid Megatron context and tensor parallelism.

\newpage

\bibliography{iclr2024_conference}

\begin{thebibliography}{81}
\providecommand{\natexlab}[1]{#1}
\providecommand{\url}[1]{\texttt{#1}}
\expandafter\ifx\csname urlstyle\endcsname\relax
  \providecommand{\doi}[1]{doi: #1}\else
  \providecommand{\doi}{doi: \begingroup \urlstyle{rm}\Url}\fi

\bibitem[qwe(2024)]{qwen2}
Qwen2 technical report.
\newblock 2024.

\bibitem[Bai et~al.(2023)Bai, Bai, Yang, Wang, Tan, Wang, Lin, Zhou, and Zhou]{bai2023qwen}
Jinze Bai, Shuai Bai, Shusheng Yang, Shijie Wang, Sinan Tan, Peng Wang, Junyang Lin, Chang Zhou, and Jingren Zhou.
\newblock Qwen-vl: A frontier large vision-language model with versatile abilities.
\newblock \emph{arXiv preprint arXiv:2308.12966}, 2023.

\bibitem[Bavishi et~al.(2023)Bavishi, Elsen, Hawthorne, Nye, Odena, Somani, and Ta\c{s}\i{}rlar]{fuyu-8b}
Rohan Bavishi, Erich Elsen, Curtis Hawthorne, Maxwell Nye, Augustus Odena, Arushi Somani, and Sa\u{g}nak Ta\c{s}\i{}rlar.
\newblock Introducing our multimodal models, 2023.

\bibitem[Betker et~al.(2023)Betker, Goh, Jing, Brooks, Wang, Li, Ouyang, Zhuang, Lee, Guo, et~al.]{betker2023improving}
James Betker, Gabriel Goh, Li~Jing, Tim Brooks, Jianfeng Wang, Linjie Li, Long Ouyang, Juntang Zhuang, Joyce Lee, Yufei Guo, et~al.
\newblock Improving image generation with better captions.
\newblock \emph{Computer Science. https://cdn. openai. com/papers/dall-e-3. pdf}, 2\penalty0 (3):\penalty0 8, 2023.

\bibitem[Brohan et~al.(2022)Brohan, Brown, Carbajal, Chebotar, Dabis, Finn, Gopalakrishnan, Hausman, Herzog, Hsu, et~al.]{brohan2022rt}
Anthony Brohan, Noah Brown, Justice Carbajal, Yevgen Chebotar, Joseph Dabis, Chelsea Finn, Keerthana Gopalakrishnan, Karol Hausman, Alex Herzog, Jasmine Hsu, et~al.
\newblock Rt-1: Robotics transformer for real-world control at scale.
\newblock \emph{arXiv preprint arXiv:2212.06817}, 2022.

\bibitem[Brohan et~al.(2023)Brohan, Brown, Carbajal, Chebotar, Chen, Choromanski, Ding, Driess, Dubey, Finn, et~al.]{brohan2023rt}
Anthony Brohan, Noah Brown, Justice Carbajal, Yevgen Chebotar, Xi~Chen, Krzysztof Choromanski, Tianli Ding, Danny Driess, Avinava Dubey, Chelsea Finn, et~al.
\newblock Rt-2: Vision-language-action models transfer web knowledge to robotic control.
\newblock \emph{arXiv preprint arXiv:2307.15818}, 2023.

\bibitem[Brown et~al.(2020)Brown, Mann, Ryder, Subbiah, Kaplan, Dhariwal, Neelakantan, Shyam, Sastry, Askell, et~al.]{brown2020language}
Tom Brown, Benjamin Mann, Nick Ryder, Melanie Subbiah, Jared~D Kaplan, Prafulla Dhariwal, Arvind Neelakantan, Pranav Shyam, Girish Sastry, Amanda Askell, et~al.
\newblock Language models are few-shot learners.
\newblock \emph{Advances in neural information processing systems}, 33:\penalty0 1877--1901, 2020.

\bibitem[Byeon et~al.(2022)Byeon, Park, Kim, Lee, Baek, and Kim]{kakaobrain2022coyo-700m}
Minwoo Byeon, Beomhee Park, Haecheon Kim, Sungjun Lee, Woonhyuk Baek, and Saehoon Kim.
\newblock Coyo-700m: Image-text pair dataset, 2022.

\bibitem[Chen et~al.(2024{\natexlab{a}})Chen, Wei, Li, Dong, Zhang, Zang, Chen, Duan, Lin, Tang, et~al.]{chen2024sharegpt4video}
Lin Chen, Xilin Wei, Jinsong Li, Xiaoyi Dong, Pan Zhang, Yuhang Zang, Zehui Chen, Haodong Duan, Bin Lin, Zhenyu Tang, et~al.
\newblock Sharegpt4video: Improving video understanding and generation with better captions.
\newblock \emph{arXiv preprint arXiv:2406.04325}, 2024{\natexlab{a}}.

\bibitem[Chen et~al.(2023)Chen, Qian, Tang, Lai, Liu, Han, and Jia]{chen2023longlora}
Yukang Chen, Shengju Qian, Haotian Tang, Xin Lai, Zhijian Liu, Song Han, and Jiaya Jia.
\newblock Longlora: Efficient fine-tuning of long-context large language models.
\newblock \emph{arXiv preprint arXiv:2309.12307}, 2023.

\bibitem[Chen et~al.(2024{\natexlab{b}})Chen, Qian, Tang, Lai, Liu, Han, and Jia]{longlora}
Yukang Chen, Shengju Qian, Haotian Tang, Xin Lai, Zhijian Liu, Song Han, and Jiaya Jia.
\newblock Longlora: Efficient fine-tuning of long-context large language models.
\newblock In \emph{The International Conference on Learning Representations}, 2024{\natexlab{b}}.

\bibitem[Chen et~al.(2024{\natexlab{c}})Chen, Wang, Tian, Ye, Gao, Cui, Tong, Hu, Luo, Ma, Ma, Wang, Dong, Yan, Guo, He, Shi, Jin, Xu, Wang, Wei, Li, Zhang, Zhang, Cai, Wen, Yan, Dou, Lu, Zhu, Lu, Lin, Qiao, Dai, and Wang]{intervl}
Zhe Chen, Weiyun Wang, Hao Tian, Shenglong Ye, Zhangwei Gao, Erfei Cui, Wenwen Tong, Kongzhi Hu, Jiapeng Luo, Zheng Ma, Ji~Ma, Jiaqi Wang, Xiaoyi Dong, Hang Yan, Hewei Guo, Conghui He, Botian Shi, Zhenjiang Jin, Chao Xu, Bin Wang, Xingjian Wei, Wei Li, Wenjian Zhang, Bo~Zhang, Pinlong Cai, Licheng Wen, Xiangchao Yan, Min Dou, Lewei Lu, Xizhou Zhu, Tong Lu, Dahua Lin, Yu~Qiao, Jifeng Dai, and Wenhai Wang.
\newblock How far are we to gpt-4v? closing the gap to commercial multimodal models with open-source suites.
\newblock \emph{CoRR}, abs/2404.16821, 2024{\natexlab{c}}.

\bibitem[Cheng et~al.(2024)Cheng, Leng, Zhang, Xin, Li, Chen, Zhu, Zhang, Luo, Zhao, and Bing]{videollama2}
Zesen Cheng, Sicong Leng, Hang Zhang, Yifei Xin, Xin Li, Guanzheng Chen, Yongxin Zhu, Wenqi Zhang, Ziyang Luo, Deli Zhao, and Lidong Bing.
\newblock Videollama 2: Advancing spatial-temporal modeling and audio understanding in video-llms.
\newblock \emph{CoRR}, abs/2406.07476, 2024.

\bibitem[Chowdhery et~al.(2023)Chowdhery, Narang, Devlin, Bosma, Mishra, Roberts, Barham, Chung, Sutton, Gehrmann, et~al.]{chowdhery2023palm}
Aakanksha Chowdhery, Sharan Narang, Jacob Devlin, Maarten Bosma, Gaurav Mishra, Adam Roberts, Paul Barham, Hyung~Won Chung, Charles Sutton, Sebastian Gehrmann, et~al.
\newblock Palm: Scaling language modeling with pathways.
\newblock \emph{Journal of Machine Learning Research}, 24\penalty0 (240):\penalty0 1--113, 2023.

\bibitem[Dai et~al.(2023)Dai, Li, Li, Tiong, Zhao, Wang, Li, Fung, and Hoi]{Dai2023InstructBLIP}
Wenliang Dai, Junnan Li, Dongxu Li, Anthony Meng~Huat Tiong, Junqi Zhao, Weisheng Wang, Boyang~Albert Li, Pascale Fung, and Steven C.~H. Hoi.
\newblock Instructblip: Towards general-purpose vision-language models with instruction tuning.
\newblock \emph{ArXiv}, abs/2305.06500, 2023.

\bibitem[Dao(2024)]{flash-attention2}
Tri Dao.
\newblock Flashattention-2: Faster attention with better parallelism and work partitioning.
\newblock In \emph{ICLR}, 2024.

\bibitem[Dehghani et~al.(2023)Dehghani, Djolonga, Mustafa, Padlewski, Heek, Gilmer, Steiner, Caron, Geirhos, Alabdulmohsin, et~al.]{dehghani2023scaling}
Mostafa Dehghani, Josip Djolonga, Basil Mustafa, Piotr Padlewski, Jonathan Heek, Justin Gilmer, Andreas~Peter Steiner, Mathilde Caron, Robert Geirhos, Ibrahim Alabdulmohsin, et~al.
\newblock Scaling vision transformers to 22 billion parameters.
\newblock In \emph{International Conference on Machine Learning}, pp.\  7480--7512. PMLR, 2023.

\bibitem[Driess et~al.(2023)Driess, Xia, Sajjadi, Lynch, Chowdhery, Ichter, Wahid, Tompson, Vuong, Yu, et~al.]{driess2023palm}
Danny Driess, Fei Xia, Mehdi~SM Sajjadi, Corey Lynch, Aakanksha Chowdhery, Brian Ichter, Ayzaan Wahid, Jonathan Tompson, Quan Vuong, Tianhe Yu, et~al.
\newblock Palm-e: An embodied multimodal language model.
\newblock \emph{arXiv preprint arXiv:2303.03378}, 2023.

\bibitem[Du et~al.(2024)Du, Zhou, Huo, Li, Zhao, Lu, Zhao, Wang, Chen, and Wen]{eventbench}
Yifan Du, Kun Zhou, Yuqi Huo, Yifan Li, Wayne~Xin Zhao, Haoyu Lu, Zijia Zhao, Bingning Wang, Weipeng Chen, and Ji{-}Rong Wen.
\newblock Towards event-oriented long video understanding.
\newblock \emph{CoRR}, abs/2406.14129, 2024.

\bibitem[Fang \& Zhao(2024)Fang and Zhao]{USP}
Jiarui Fang and Shangchun Zhao.
\newblock Usp: A unified sequence parallelism approach for long context generative ai.
\newblock \emph{arXiv preprint arXiv:2405.07719}, 2024.

\bibitem[Fang et~al.(2024)Fang, Zhu, Lu, Wang, Molchanov, Cho, Pavone, Han, and Yin]{fang2024vila}
Yunhao Fang, Ligeng Zhu, Yao Lu, Yan Wang, Pavlo Molchanov, Jang~Hyun Cho, Marco Pavone, Song Han, and Hongxu Yin.
\newblock Vila$^2$: Vila augmented vila.
\newblock \emph{arXiv preprint arXiv:2407.17453}, 2024.

\bibitem[Fei et~al.(2024)Fei, Li, Deng, Wang, Liu, and Wang]{video-ccam}
Jiajun Fei, Dian Li, Zhidong Deng, Zekun Wang, Gang Liu, and Hui Wang.
\newblock Video-ccam: Enhancing video-language understanding with causal cross-attention masks for short and long videos.
\newblock \emph{CoRR}, abs/2408.14023, 2024.

\bibitem[Fu et~al.(2024{\natexlab{a}})Fu, Dai, Luo, Li, Ren, Zhang, Wang, Zhou, Shen, Zhang, Chen, Li, Lin, Zhao, Li, Xu, Zheng, Chen, Ji, and Sun]{video-mme}
Chaoyou Fu, Yuhan Dai, Yondong Luo, Lei Li, Shuhuai Ren, Renrui Zhang, Zihan Wang, Chenyu Zhou, Yunhang Shen, Mengdan Zhang, Peixian Chen, Yanwei Li, Shaohui Lin, Sirui Zhao, Ke~Li, Tong Xu, Xiawu Zheng, Enhong Chen, Rongrong Ji, and Xing Sun.
\newblock Video-mme: The first-ever comprehensive evaluation benchmark of multi-modal llms in video analysis.
\newblock \emph{CoRR}, abs/2405.21075, 2024{\natexlab{a}}.

\bibitem[Fu et~al.(2024{\natexlab{b}})Fu, Lin, Long, Shen, Zhao, Zhang, Wang, Yin, Ma, Zheng, He, Ji, Wu, Shan, and Sun]{vita}
Chaoyou Fu, Haojia Lin, Zuwei Long, Yunhang Shen, Meng Zhao, Yifan Zhang, Xiong Wang, Di~Yin, Long Ma, Xiawu Zheng, Ran He, Rongrong Ji, Yunsheng Wu, Caifeng Shan, and Xing Sun.
\newblock {VITA:} towards open-source interactive omni multimodal {LLM}.
\newblock \emph{CoRR}, abs/2408.05211, 2024{\natexlab{b}}.

\bibitem[Fu et~al.(2024{\natexlab{c}})Fu, Panda, Niu, Yue, Hajishirzi, Kim, and Peng]{fu2024data}
Yao Fu, Rameswar Panda, Xinyao Niu, Xiang Yue, Hannaneh Hajishirzi, Yoon Kim, and Hao Peng.
\newblock Data engineering for scaling language models to 128k context.
\newblock \emph{arXiv preprint arXiv:2402.10171}, 2024{\natexlab{c}}.

\bibitem[Fu et~al.(2024{\natexlab{d}})Fu, Panda, Niu, Yue, Hajishirzi, Kim, and Peng]{long-context-data-engineering}
Yao Fu, Rameswar Panda, Xinyao Niu, Xiang Yue, Hannaneh Hajishirzi, Yoon Kim, and Hao Peng.
\newblock Data engineering for scaling language models to 128k context.
\newblock \emph{CoRR}, abs/2402.10171, 2024{\natexlab{d}}.

\bibitem[Gu et~al.(2024)Gu, Sun, Hu, Huang, Chen, Xiong, Wang, Chen, Zhao, Fang, Wen, Zhang, Jin, and Liu]{LoongTrain}
Diandian Gu, Peng Sun, Qinghao Hu, Ting Huang, Xun Chen, Yingtong Xiong, Guoteng Wang, Qiaoling Chen, Shangchun Zhao, Jiarui Fang, Yonggang Wen, Tianwei Zhang, Xin Jin, and Xuanzhe Liu.
\newblock Loongtrain: Efficient training of long-sequence llms with head-context parallelism.
\newblock \emph{CoRR}, pdf/2406.18485, 2024.

\bibitem[Han et~al.(2023)Han, Yang, Chang, and Wang]{han2023shot2story20k}
Mingfei Han, Linjie Yang, Xiaojun Chang, and Heng Wang.
\newblock Shot2story20k: A new benchmark for comprehensive understanding of multi-shot videos.
\newblock \emph{arXiv preprint arXiv:2311.17043}, 2023.

\bibitem[Huang et~al.(2019)Huang, Cheng, Bapna, Firat, Chen, Chen, Lee, Ngiam, Le, Wu, et~al.]{huang2019gpipe}
Yanping Huang, Youlong Cheng, Ankur Bapna, Orhan Firat, Dehao Chen, Mia Chen, HyoukJoong Lee, Jiquan Ngiam, Quoc~V Le, Yonghui Wu, et~al.
\newblock Gpipe: Efficient training of giant neural networks using pipeline parallelism.
\newblock \emph{Advances in neural information processing systems}, 32, 2019.

\bibitem[Jacobs et~al.(2023)Jacobs, Tanaka, Zhang, Zhang, Song, Rajbhandari, and He]{jacobs2023deepspeed}
Sam~Ade Jacobs, Masahiro Tanaka, Chengming Zhang, Minjia Zhang, Leon Song, Samyam Rajbhandari, and Yuxiong He.
\newblock Deepspeed ulysses: System optimizations for enabling training of extreme long sequence transformer models.
\newblock \emph{arXiv preprint arXiv:2309.14509}, 2023.

\bibitem[Jin et~al.(2023)Jin, Takanobu, Zhang, Cao, and Yuan]{chat-univi}
Peng Jin, Ryuichi Takanobu, Caiwan Zhang, Xiaochun Cao, and Li~Yuan.
\newblock Chat-univi: Unified visual representation empowers large language models with image and video understanding.
\newblock \emph{CoRR}, abs/2311.08046, 2023.

\bibitem[Koh et~al.(2024)Koh, Lo, Jang, Duvvur, Lim, Huang, Neubig, Zhou, Salakhutdinov, and Fried]{koh2024visualwebarena}
Jing~Yu Koh, Robert Lo, Lawrence Jang, Vikram Duvvur, Ming~Chong Lim, Po-Yu Huang, Graham Neubig, Shuyan Zhou, Ruslan Salakhutdinov, and Daniel Fried.
\newblock Visualwebarena: Evaluating multimodal agents on realistic visual web tasks.
\newblock \emph{arXiv preprint arXiv:2401.13649}, 2024.

\bibitem[Korthikanti et~al.(2023)Korthikanti, Casper, Lym, McAfee, Andersch, Shoeybi, and Catanzaro]{korthikanti2023reducing}
Vijay~Anand Korthikanti, Jared Casper, Sangkug Lym, Lawrence McAfee, Michael Andersch, Mohammad Shoeybi, and Bryan Catanzaro.
\newblock Reducing activation recomputation in large transformer models.
\newblock \emph{Proceedings of Machine Learning and Systems}, 5:\penalty0 341--353, 2023.

\bibitem[Lepikhin et~al.(2020)Lepikhin, Lee, Xu, Chen, Firat, Huang, Krikun, Shazeer, and Chen]{lepikhin2020gshard}
Dmitry Lepikhin, HyoukJoong Lee, Yuanzhong Xu, Dehao Chen, Orhan Firat, Yanping Huang, Maxim Krikun, Noam Shazeer, and Zhifeng Chen.
\newblock Gshard: Scaling giant models with conditional computation and automatic sharding.
\newblock \emph{arXiv preprint arXiv:2006.16668}, 2020.

\bibitem[Li et~al.(2024{\natexlab{a}})Li, Zhang, Guo, Zhang, Li, Zhang, Zhang, Li, Liu, and Li]{llava-onevision}
Bo~Li, Yuanhan Zhang, Dong Guo, Renrui Zhang, Feng Li, Hao Zhang, Kaichen Zhang, Yanwei Li, Ziwei Liu, and Chunyuan Li.
\newblock Llava-onevision: Easy visual task transfer.
\newblock \emph{CoRR}, abs/2408.03326, 2024{\natexlab{a}}.

\bibitem[Li et~al.(2023{\natexlab{a}})Li, Shao, Xie, Xing, Gonzalez, Stoica, Ma, and Zhang]{li2023lightseq}
Dacheng Li, Rulin Shao, Anze Xie, Eric Xing, Joseph Gonzalez, Ion Stoica, Xuezhe Ma, and Hao Zhang.
\newblock Lightseq: : Sequence level parallelism for distributed training of long context transformers.
\newblock In \emph{Workshop on Advancing Neural Network Training: Computational Efficiency, Scalability, and Resource Optimization}, 2023{\natexlab{a}}.

\bibitem[Li et~al.(2023{\natexlab{b}})Li, Li, Savarese, and Hoi]{li2023blip}
Junnan Li, Dongxu Li, Silvio Savarese, and Steven Hoi.
\newblock Blip-2: Bootstrapping language-image pre-training with frozen image encoders and large language models.
\newblock \emph{arXiv preprint arXiv:2301.12597}, 2023{\natexlab{b}}.

\bibitem[Li et~al.(2023{\natexlab{c}})Li, He, Wang, Li, Wang, Luo, Wang, Wang, and Qiao]{videochat}
Kunchang Li, Yinan He, Yi~Wang, Yizhuo Li, Wenhai Wang, Ping Luo, Yali Wang, Limin Wang, and Yu~Qiao.
\newblock Videochat: Chat-centric video understanding.
\newblock \emph{CoRR}, abs/2305.06355, 2023{\natexlab{c}}.

\bibitem[Li et~al.(2024{\natexlab{b}})Li, Wang, He, Li, Wang, Liu, Wang, Xu, Chen, Lou, Wang, and Qiao]{mvbench}
Kunchang Li, Yali Wang, Yinan He, Yizhuo Li, Yi~Wang, Yi~Liu, Zun Wang, Jilan Xu, Guo Chen, Ping Lou, Limin Wang, and Yu~Qiao.
\newblock Mvbench: {A} comprehensive multi-modal video understanding benchmark.
\newblock In \emph{CVPR}, pp.\  22195--22206, 2024{\natexlab{b}}.

\bibitem[Li et~al.(2021)Li, Xue, Baranwal, Li, and You]{li2021sequence}
Shenggui Li, Fuzhao Xue, Chaitanya Baranwal, Yongbin Li, and Yang You.
\newblock Sequence parallelism: Long sequence training from system perspective.
\newblock \emph{arXiv preprint arXiv:2105.13120}, 2021.

\bibitem[Li et~al.(2024{\natexlab{c}})Li, Zhang, Wang, Zhong, Chen, Chu, Liu, and Jia]{li2024minigeminiminingpotentialmultimodality}
Yanwei Li, Yuechen Zhang, Chengyao Wang, Zhisheng Zhong, Yixin Chen, Ruihang Chu, Shaoteng Liu, and Jiaya Jia.
\newblock Mini-gemini: Mining the potential of multi-modality vision language models, 2024{\natexlab{c}}.

\bibitem[Lin et~al.(2023{\natexlab{a}})Lin, Ye, Zhu, Cui, Ning, Jin, and Yuan]{video-llava}
Bin Lin, Yang Ye, Bin Zhu, Jiaxi Cui, Munan Ning, Peng Jin, and Li~Yuan.
\newblock Video-llava: Learning united visual representation by alignment before projection.
\newblock \emph{CoRR}, abs/2311.10122, 2023{\natexlab{a}}.

\bibitem[Lin et~al.(2023{\natexlab{b}})Lin, Yin, Ping, Lu, Molchanov, Tao, Mao, Kautz, Shoeybi, and Han]{lin2023vila}
Ji~Lin, Hongxu Yin, Wei Ping, Yao Lu, Pavlo Molchanov, Andrew Tao, Huizi Mao, Jan Kautz, Mohammad Shoeybi, and Song Han.
\newblock Vila: On pre-training for visual language models, 2023{\natexlab{b}}.

\bibitem[Liu et~al.(2023{\natexlab{a}})Liu, Zaharia, and Abbeel]{liu2023ring}
Hao Liu, Matei Zaharia, and Pieter Abbeel.
\newblock Ring attention with blockwise transformers for near-infinite context.
\newblock \emph{arXiv preprint arXiv:2310.01889}, 2023{\natexlab{a}}.

\bibitem[Liu et~al.(2024{\natexlab{a}})Liu, Yan, Zaharia, and Abbeel]{liu2024world}
Hao Liu, Wilson Yan, Matei Zaharia, and Pieter Abbeel.
\newblock World model on million-length video and language with blockwise ringattention.
\newblock \emph{arXiv preprint arXiv:2402.08268}, 2024{\natexlab{a}}.

\bibitem[Liu et~al.(2023{\natexlab{b}})Liu, Li, Li, and Lee]{liu2023improved}
Haotian Liu, Chunyuan Li, Yuheng Li, and Yong~Jae Lee.
\newblock Improved baselines with visual instruction tuning.
\newblock \emph{arXiv preprint arXiv:2310.03744}, 2023{\natexlab{b}}.

\bibitem[Liu et~al.(2023{\natexlab{c}})Liu, Li, Wu, and Lee]{liu2023llava}
Haotian Liu, Chunyuan Li, Qingyang Wu, and Yong~Jae Lee.
\newblock Visual instruction tuning.
\newblock In \emph{NeurIPS}, 2023{\natexlab{c}}.

\bibitem[Liu et~al.(2024{\natexlab{b}})Liu, Li, Li, Li, Zhang, Shen, and Lee]{liu2024llavanext}
Haotian Liu, Chunyuan Li, Yuheng Li, Bo~Li, Yuanhan Zhang, Sheng Shen, and Yong~Jae Lee.
\newblock Llava-next: Improved reasoning, ocr, and world knowledge, January 2024{\natexlab{b}}.

\bibitem[Liu et~al.(2024{\natexlab{c}})Liu, Wang, Ma, Wu, Ma, Wei, Jiao, Wu, and Hu]{kangaroo}
Jiajun Liu, Yibing Wang, Hanghang Ma, Xiaoping Wu, Xiaoqi Ma, xiaoming Wei, Jianbin Jiao, Enhua Wu, and Jie Hu.
\newblock Kangaroo: A powerful video-language model supporting long-context video input.
\newblock \emph{arXiv preprint arXiv:2408.15542}, 2024{\natexlab{c}}.

\bibitem[Maaz et~al.(2024)Maaz, Rasheed, Khan, and Khan]{Maaz2023VideoChatGPT}
Muhammad Maaz, Hanoona Rasheed, Salman Khan, and Fahad~Shahbaz Khan.
\newblock Video-chatgpt: Towards detailed video understanding via large vision and language models.
\newblock In \emph{Proceedings of the 62nd Annual Meeting of the Association for Computational Linguistics (ACL 2024)}, 2024.

\bibitem[Mangalam et~al.(2023)Mangalam, Akshulakov, and Malik]{egoschema}
Karttikeya Mangalam, Raiymbek Akshulakov, and Jitendra Malik.
\newblock Egoschema: {A} diagnostic benchmark for very long-form video language understanding.
\newblock In \emph{NeurIPS}, 2023.

\bibitem[Narayanan et~al.(2019)Narayanan, Harlap, Phanishayee, Seshadri, Devanur, Ganger, Gibbons, and Zaharia]{narayanan2019pipedream}
Deepak Narayanan, Aaron Harlap, Amar Phanishayee, Vivek Seshadri, Nikhil~R Devanur, Gregory~R Ganger, Phillip~B Gibbons, and Matei Zaharia.
\newblock Pipedream: Generalized pipeline parallelism for dnn training.
\newblock In \emph{Proceedings of the 27th ACM symposium on operating systems principles}, pp.\  1--15, 2019.

\bibitem[Ouyang et~al.(2022)Ouyang, Wu, Jiang, Almeida, Wainwright, Mishkin, Zhang, Agarwal, Slama, Ray, et~al.]{ouyang2022training}
Long Ouyang, Jeffrey Wu, Xu~Jiang, Diogo Almeida, Carroll Wainwright, Pamela Mishkin, Chong Zhang, Sandhini Agarwal, Katarina Slama, Alex Ray, et~al.
\newblock Training language models to follow instructions with human feedback.
\newblock \emph{Advances in neural information processing systems}, 35:\penalty0 27730--27744, 2022.

\bibitem[Padalkar et~al.(2023)Padalkar, Pooley, Jain, Bewley, Herzog, Irpan, Khazatsky, Rai, Singh, Brohan, et~al.]{padalkar2023open}
Abhishek Padalkar, Acorn Pooley, Ajinkya Jain, Alex Bewley, Alex Herzog, Alex Irpan, Alexander Khazatsky, Anant Rai, Anikait Singh, Anthony Brohan, et~al.
\newblock Open x-embodiment: Robotic learning datasets and rt-x models.
\newblock \emph{arXiv preprint arXiv:2310.08864}, 2023.

\bibitem[Patraucean et~al.(2023)Patraucean, Smaira, Gupta, Recasens, Markeeva, Banarse, Koppula, Heyward, Malinowski, Yang, Doersch, Matejovicova, Sulsky, Miech, Fr{\'{e}}chette, Klimczak, Koster, Zhang, Winkler, Aytar, Osindero, Damen, Zisserman, and Carreira]{perpceptiontest}
Viorica Patraucean, Lucas Smaira, Ankush Gupta, Adri{\`{a}} Recasens, Larisa Markeeva, Dylan Banarse, Skanda Koppula, Joseph Heyward, Mateusz Malinowski, Yi~Yang, Carl Doersch, Tatiana Matejovicova, Yury Sulsky, Antoine Miech, Alexandre Fr{\'{e}}chette, Hanna Klimczak, Raphael Koster, Junlin Zhang, Stephanie Winkler, Yusuf Aytar, Simon Osindero, Dima Damen, Andrew Zisserman, and Jo{\~{a}}o Carreira.
\newblock Perception test: {A} diagnostic benchmark for multimodal video models.
\newblock In \emph{NeurIPS}, 2023.

\bibitem[Rajbhandari et~al.(2020)Rajbhandari, Rasley, Ruwase, and He]{rajbhandari2020zero}
Samyam Rajbhandari, Jeff Rasley, Olatunji Ruwase, and Yuxiong He.
\newblock Zero: Memory optimizations toward training trillion parameter models.
\newblock In \emph{SC20: International Conference for High Performance Computing, Networking, Storage and Analysis}, pp.\  1--16. IEEE, 2020.

\bibitem[Share(2024)]{sharegemini}
Share.
\newblock Sharegemini: Scaling up video caption data for multimodal large language models, June 2024.

\bibitem[Shoeybi et~al.(2019)Shoeybi, Patwary, Puri, LeGresley, Casper, and Catanzaro]{shoeybi2019megatron}
Mohammad Shoeybi, Mostofa Patwary, Raul Puri, Patrick LeGresley, Jared Casper, and Bryan Catanzaro.
\newblock Megatron-lm: Training multi-billion parameter language models using model parallelism.
\newblock \emph{arXiv preprint arXiv:1909.08053}, 2019.

\bibitem[Soboleva et~al.(2023)Soboleva, Al-Khateeb, Myers, Steeves, Hestness, and Dey]{cerebras2023slimpajama}
Daria Soboleva, Faisal Al-Khateeb, Robert Myers, Jacob~R Steeves, Joel Hestness, and Nolan Dey.
\newblock {SlimPajama: A 627B token cleaned and deduplicated version of RedPajama}, 2023.

\bibitem[Su et~al.(2021)Su, Lu, Pan, Wen, and Liu]{rope}
Jianlin Su, Yu~Lu, Shengfeng Pan, Bo~Wen, and Yunfeng Liu.
\newblock Roformer: Enhanced transformer with rotary position embedding.
\newblock \emph{CoRR}, abs/2104.09864, 2021.

\bibitem[Team(2024)]{team2024chameleon}
Chameleon Team.
\newblock Chameleon: Mixed-modal early-fusion foundation models.
\newblock \emph{arXiv preprint arXiv:2405.09818}, 2024.

\bibitem[Tillet et~al.(2019)Tillet, Kung, and Cox]{tillet2019triton}
Philippe Tillet, Hsiang-Tsung Kung, and David Cox.
\newblock Triton: an intermediate language and compiler for tiled neural network computations.
\newblock In \emph{Proceedings of the 3rd ACM SIGPLAN International Workshop on Machine Learning and Programming Languages}, pp.\  10--19, 2019.

\bibitem[Tong et~al.(2024)Tong, Brown, Wu, Woo, Middepogu, Akula, Yang, Yang, Iyer, Pan, Wang, Fergus, LeCun, and Xie]{tong2024cambrian1fullyopenvisioncentric}
Shengbang Tong, Ellis Brown, Penghao Wu, Sanghyun Woo, Manoj Middepogu, Sai~Charitha Akula, Jihan Yang, Shusheng Yang, Adithya Iyer, Xichen Pan, Austin Wang, Rob Fergus, Yann LeCun, and Saining Xie.
\newblock Cambrian-1: A fully open, vision-centric exploration of multimodal llms, 2024.

\bibitem[Weng et~al.(2024)Weng, Han, He, Chang, and Zhuang]{longvlm}
Yuetian Weng, Mingfei Han, Haoyu He, Xiaojun Chang, and Bohan Zhuang.
\newblock Longvlm: Efficient long video understanding via large language models.
\newblock \emph{CoRR}, abs/2404.03384, 2024.

\bibitem[Wu et~al.(2024)Wu, Li, Chen, and Li]{longvideobench}
Haoning Wu, Dongxu Li, Bei Chen, and Junnan Li.
\newblock Longvideobench: {A} benchmark for long-context interleaved video-language understanding.
\newblock \emph{CoRR}, abs/2407.15754, 2024.

\bibitem[Xiao et~al.(2021)Xiao, Shang, Yao, and Chua]{next-qa}
Junbin Xiao, Xindi Shang, Angela Yao, and Tat{-}Seng Chua.
\newblock Next-qa: Next phase of question-answering to explaining temporal actions.
\newblock In \emph{CVPR}, pp.\  9777--9786, 2021.

\bibitem[Xu et~al.(2024)Xu, Zhao, Zhou, Lin, Ng, and Feng]{pllava}
Lin Xu, Yilin Zhao, Daquan Zhou, Zhijie Lin, See{-}Kiong Ng, and Jiashi Feng.
\newblock Pllava : Parameter-free llava extension from images to videos for video dense captioning.
\newblock \emph{CoRR}, abs/2404.16994, 2024.

\bibitem[Ye et~al.(2024)Ye, Huang, Lu, Yu, Ping, Tao, Kautz, Han, Xu, Molchanov, and Yin]{x-vila}
Hanrong Ye, De{-}An Huang, Yao Lu, Zhiding Yu, Wei Ping, Andrew Tao, Jan Kautz, Song Han, Dan Xu, Pavlo Molchanov, and Hongxu Yin.
\newblock {X-VILA:} cross-modality alignment for large language model.
\newblock \emph{CoRR}, abs/2405.19335, 2024.

\bibitem[Yu et~al.(2022)Yu, Jeong, Kim, Kim, and Chun]{yu2022orca}
Gyeong-In Yu, Joo~Seong Jeong, Geon-Woo Kim, Soojeong Kim, and Byung-Gon Chun.
\newblock Orca: A distributed serving system for $\{$Transformer-Based$\}$ generative models.
\newblock In \emph{16th USENIX Symposium on Operating Systems Design and Implementation (OSDI 22)}, pp.\  521--538, 2022.

\bibitem[Yu et~al.(2019)Yu, Xu, Yu, Yu, Zhao, Zhuang, and Tao]{activity-qa}
Zhou Yu, Dejing Xu, Jun Yu, Ting Yu, Zhou Zhao, Yueting Zhuang, and Dacheng Tao.
\newblock Activitynet-qa: {A} dataset for understanding complex web videos via question answering.
\newblock In \emph{AAAI}, pp.\  9127--9134, 2019.

\bibitem[Zhang et~al.(2024{\natexlab{a}})Zhang, Wang, Tang, Liu, Feng, Dai, and Jin]{flash-vstream}
Haoji Zhang, Yiqin Wang, Yansong Tang, Yong Liu, Jiashi Feng, Jifeng Dai, and Xiaojie Jin.
\newblock Flash-vstream: Memory-based real-time understanding for long video streams.
\newblock \emph{CoRR}, abs/2406.08085, 2024{\natexlab{a}}.

\bibitem[Zhang et~al.()Zhang, Zhang, Li, Zeng, Yang, Zhang, Wang, Tan, Li, and Liu]{longva}
Peiyuan Zhang, Kaichen Zhang, Bo~Li, Guangtao Zeng, Jingkang Yang, Yuanhan Zhang, Ziyue Wang, Haoran Tan, Chunyuan Li, and Ziwei Liu.
\newblock Long context transfer from language to vision.
\newblock \emph{CoRR}.

\bibitem[Zhang et~al.(2024{\natexlab{b}})Zhang, Zhang, Li, Zeng, Yang, Zhang, Wang, Tan, Li, and Liu]{zhang2024longva}
Peiyuan Zhang, Kaichen Zhang, Bo~Li, Guangtao Zeng, Jingkang Yang, Yuanhan Zhang, Ziyue Wang, Haoran Tan, Chunyuan Li, and Ziwei Liu.
\newblock Long context transfer from language to vision.
\newblock \emph{arXiv preprint arXiv:2406.16852}, 2024{\natexlab{b}}.

\bibitem[Zhang et~al.(2024{\natexlab{c}})Zhang, Gui, Sun, Feng, Xu, Zhang, Fu, Li, Hauptmann, Bisk, and Yang]{zhang2024direct}
Ruohong Zhang, Liangke Gui, Zhiqing Sun, Yihao Feng, Keyang Xu, Yuanhan Zhang, Di~Fu, Chunyuan Li, Alexander Hauptmann, Yonatan Bisk, and Yiming Yang.
\newblock Direct preference optimization of video large multimodal models from language model reward, 2024{\natexlab{c}}.

\bibitem[Zhang et~al.(2024{\natexlab{d}})Zhang, Wen, Fu, Wang, Zhang, Wang, and Jin]{slime}
Yi{-}Fan Zhang, Qingsong Wen, Chaoyou Fu, Xue Wang, Zhang Zhang, Liang Wang, and Rong Jin.
\newblock Beyond llava-hd: Diving into high-resolution large multimodal models.
\newblock \emph{CoRR}, abs/2406.08487, 2024{\natexlab{d}}.

\bibitem[Zhao et~al.(2023)Zhao, Gu, Varma, Luo, Huang, Xu, Wright, Shojanazeri, Ott, Shleifer, et~al.]{zhao2023pytorch}
Yanli Zhao, Andrew Gu, Rohan Varma, Liang Luo, Chien-Chin Huang, Min Xu, Less Wright, Hamid Shojanazeri, Myle Ott, Sam Shleifer, et~al.
\newblock Pytorch fsdp: experiences on scaling fully sharded data parallel.
\newblock \emph{arXiv preprint arXiv:2304.11277}, 2023.

\bibitem[Zhao et~al.(2024)Zhao, Lu, Huo, Du, Yue, Guo, Wang, Chen, and Liu]{vnbench}
Zijia Zhao, Haoyu Lu, Yuqi Huo, Yifan Du, Tongtian Yue, Longteng Guo, Bingning Wang, Weipeng Chen, and Jing Liu.
\newblock Needle in a video haystack: A scalable synthetic framework for benchmarking video mllms.
\newblock \emph{arXiv preprint}, 2024.

\bibitem[Zheng et~al.(2023)Zheng, Chiang, Sheng, Zhuang, Wu, Zhuang, Lin, Li, Li, Xing, Zhang, Gonzalez, and Stoica]{zheng2023judging}
Lianmin Zheng, Wei-Lin Chiang, Ying Sheng, Siyuan Zhuang, Zhanghao Wu, Yonghao Zhuang, Zi~Lin, Zhuohan Li, Dacheng Li, Eric.~P Xing, Hao Zhang, Joseph~E. Gonzalez, and Ion Stoica.
\newblock Judging llm-as-a-judge with mt-bench and chatbot arena, 2023.

\bibitem[Zhou et~al.(2024)Zhou, Liu, Xu, Iyer, Sun, Mao, Ma, Efrat, Yu, Yu, et~al.]{zhou2024lima}
Chunting Zhou, Pengfei Liu, Puxin Xu, Srinivasan Iyer, Jiao Sun, Yuning Mao, Xuezhe Ma, Avia Efrat, Ping Yu, Lili Yu, et~al.
\newblock Lima: Less is more for alignment.
\newblock \emph{Advances in Neural Information Processing Systems}, 36, 2024.

\bibitem[Zhou et~al.(2018)Zhou, Xu, and Corso]{youcook2}
Luowei Zhou, Chenliang Xu, and Jason~J. Corso.
\newblock Towards automatic learning of procedures from web instructional videos.
\newblock In \emph{AAAI}, pp.\  7590--7598, 2018.

\bibitem[Zhu(2023)]{ring-flash-attention}
Zilin Zhu.
\newblock Ring flash attention, 2023.
\newblock Accessed: 2024-07-28.

\end{thebibliography}
\bibliographystyle{iclr2024_conference}

\newpage
\appendix
\section{Appendix}

\subsection{{LongVILA-Caption}}
\begin{figure}[h]
    \centering
    \includegraphics[width=\linewidth]{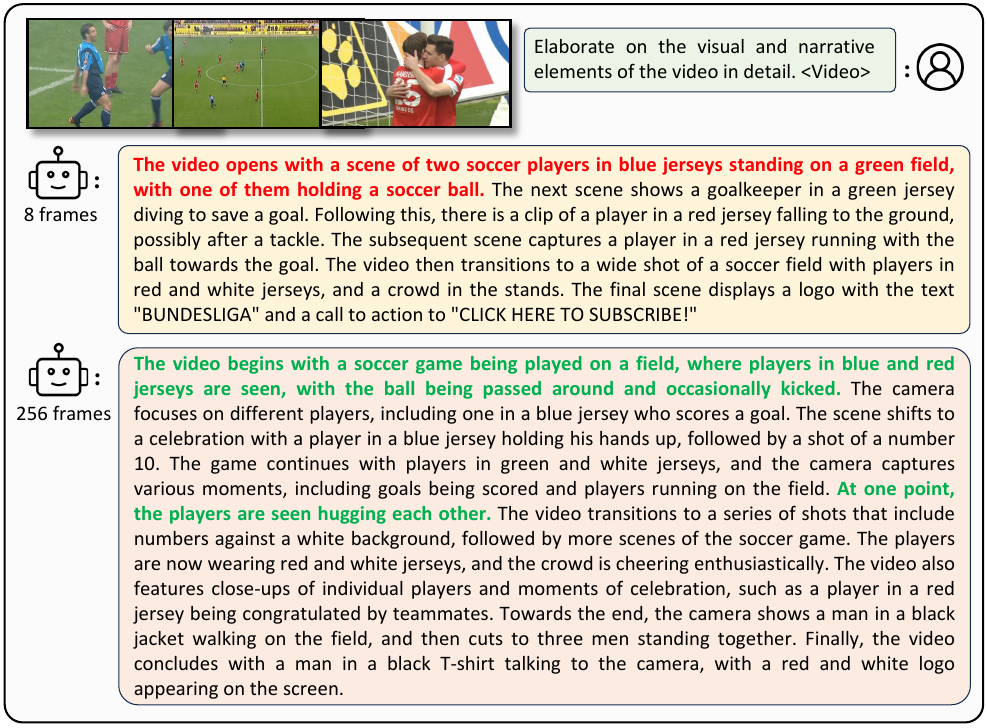}
    \caption{{Examples of sports long video caption with LongVILA. For the gameplay opening, the 8-frame baseline describes only static image, two players in only blue jerseys. In contrast, 256-frame LongVILA describes players in blue and red jerseys passing and kicking the ball. In addition, the 256-frame version also include the detail of players hugging emphasizes the celebratory aspects, which is missing in the 8-frame baseline.}}
    \label{fig:caption-demo-1}
\end{figure}

\begin{figure}[h]
    \centering
    \includegraphics[width=\linewidth]{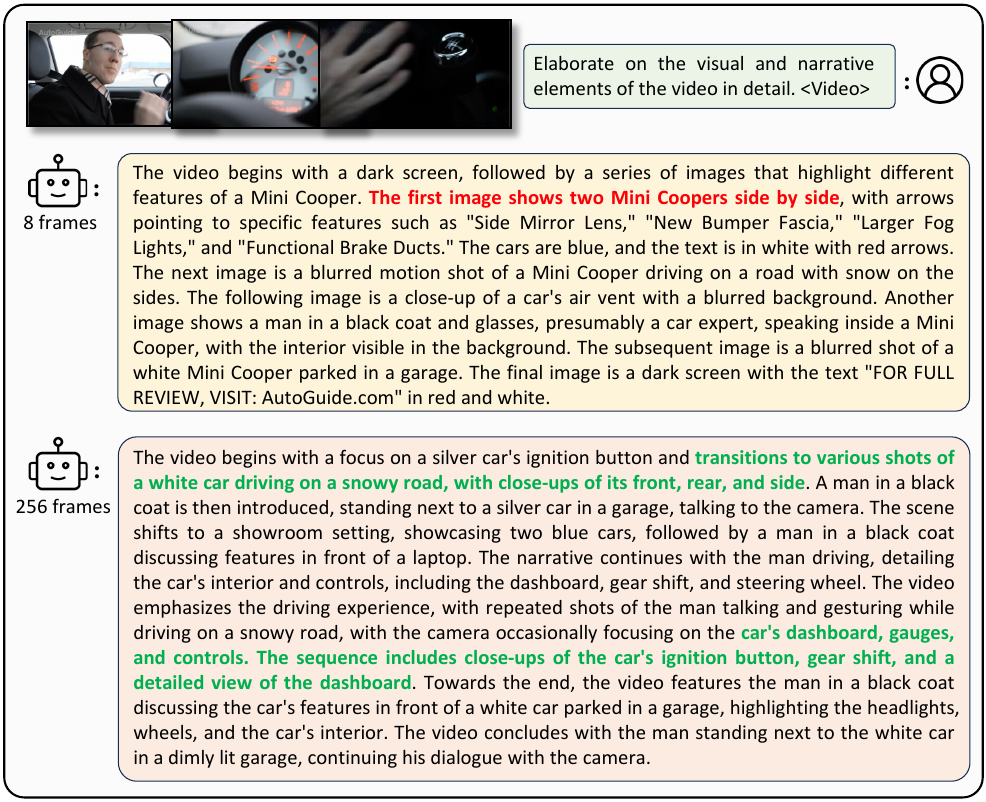}
    \caption{{Examples of technology long video caption with LongVILA. At the beginning of captions, the 8-frame baseline only describes static image and two cars. In contrast, the 256-frame LongVILA describes the car on snowy road, covering front, rear, and side views.
    For details, the 256-frame LongVILA describes close-ups of ignition button, gear shift, and dashboard elements, which are missing in the 8-frame baseline.}}
    \label{fig:caption-demo-2}
\end{figure}

{We have developed a long video captioning benchmark, LongVILA-Caption, consisting of 100 long videos, with captions generated as detailed in Section~\ref{sec:method-stage4}, and verified through human examination. In line with the methodology of VideoChatGPT \citep{Maaz2023VideoChatGPT}, we evaluate the predictions of each model based on their correctness, detailed orientation, and contextual understanding. For instance, we assess correctness by employing GPT-4 to predict scores using a specific prompt. Additionally, we present two examples in Figures~\ref{fig:caption-demo-1} and \ref{fig:caption-demo-2}, featuring long videos in sports and technology. These examples demonstrate that LongVILA, with its capability to process more frames, offers a more comprehensive understanding of videos compared to its short-frame counterpart.}

{The Table~\ref{tab:longvila-caption} presents the performance metrics for the LongVILA models being trained and evaluated on varying numbers of frames: 8, 128, and 256.  As the number of frames increases, the model's performance improves significantly. Specifically, the average scores rise from 2.00 to 3.26, highlighting the model's enhanced capability in generating accurate and rich captions with more frames.}

\begin{table}[h]
\centering
\begin{minipage}{0.42\textwidth}
\centering
\caption{Iteration time (seconds) on the dataset~\citep{chen2024sharegpt4video} with and without our two-stage sharding strategy.}
\resizebox{1.0\textwidth}{!}{\begin{tabular}{c|ccc}
\toprule     & 2 GPUs & 4 GPUs  & 8 GPUs  \\ 
\midrule
one-stage & 0.78 & 0.89 & 1.20\\
two-stage & 0.77 & 0.86 & 1.12  \\
\bottomrule
\end{tabular}
}
\label{tab:two_stage}
\end{minipage}
\hfill
\begin{minipage}{0.56\textwidth}
\centering
\caption{{Evaluation of LongVILA-Caption performance across different frame counts.}}
\resizebox{\textwidth}{!}{\begin{tabular}{c|cccc}
\toprule                                                                Frames  & Correctness  & Detailed  & Contextual  & Average \\ \midrule
8  & 1.87 & 1.85  & 2.27  & 2.00 \\ 
128  & 2.36  & 2.44  & 2.79 & 2.53   \\
256  & 3.23  & 3.11  & 3.43  & 3.26   \\ \bottomrule
\end{tabular}}
\label{tab:longvila-caption}
\end{minipage}
\end{table}

\begin{figure}[h]
    \centering
    \includegraphics[width=1.0\linewidth]{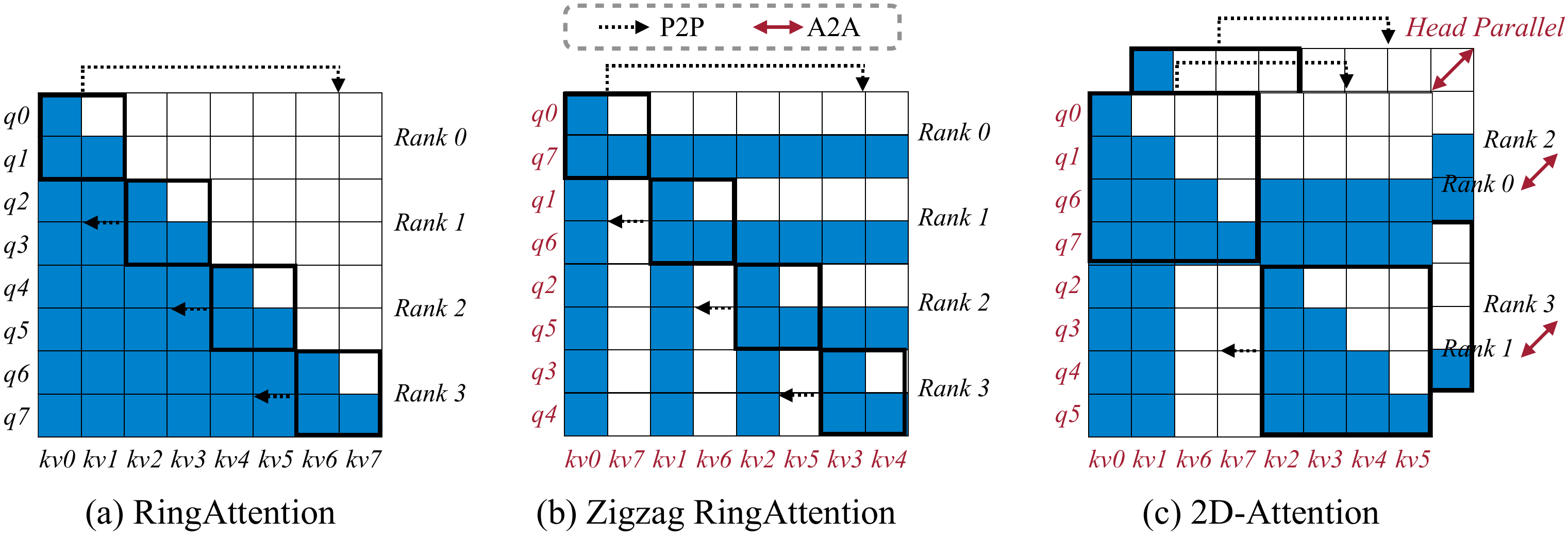}
    \caption{{Comparison of RingAttention ~\citet{liu2023ring}, \zigzag~\citet{ring-flash-attention}, and 2D-Attention \citep{USP}. The blue block indicates communication between QKV, while the black frame represents local attention computation within each SP group rank. The sequence length is 8 and the global SP degree is 4. Due to the triangular structure of causal attention computations, RingAttention experiences a computation imbalance, where rank 0 becomes idle after the first round while rank 3 continues computing through all stages. \zigzag addresses this by reordering input sequence tokens along the sequence dimension to achieve load balance. The 2D-Attention mechanism uses a ring parallel degree of 2 and a head parallel degree of 2, resulting in an effective global sequence parallel degree of 4. This approach also incorporates a workload balancing strategy within the ring-based process group and uses the All-to-All operation to distribute QKV tensors across devices based on the head dimension, ensuring efficient and balanced computation.}}
    \label{fig:system_2d_attn}

\end{figure}

\begin{figure}[t]
    \centering
    \includegraphics[width=\linewidth]{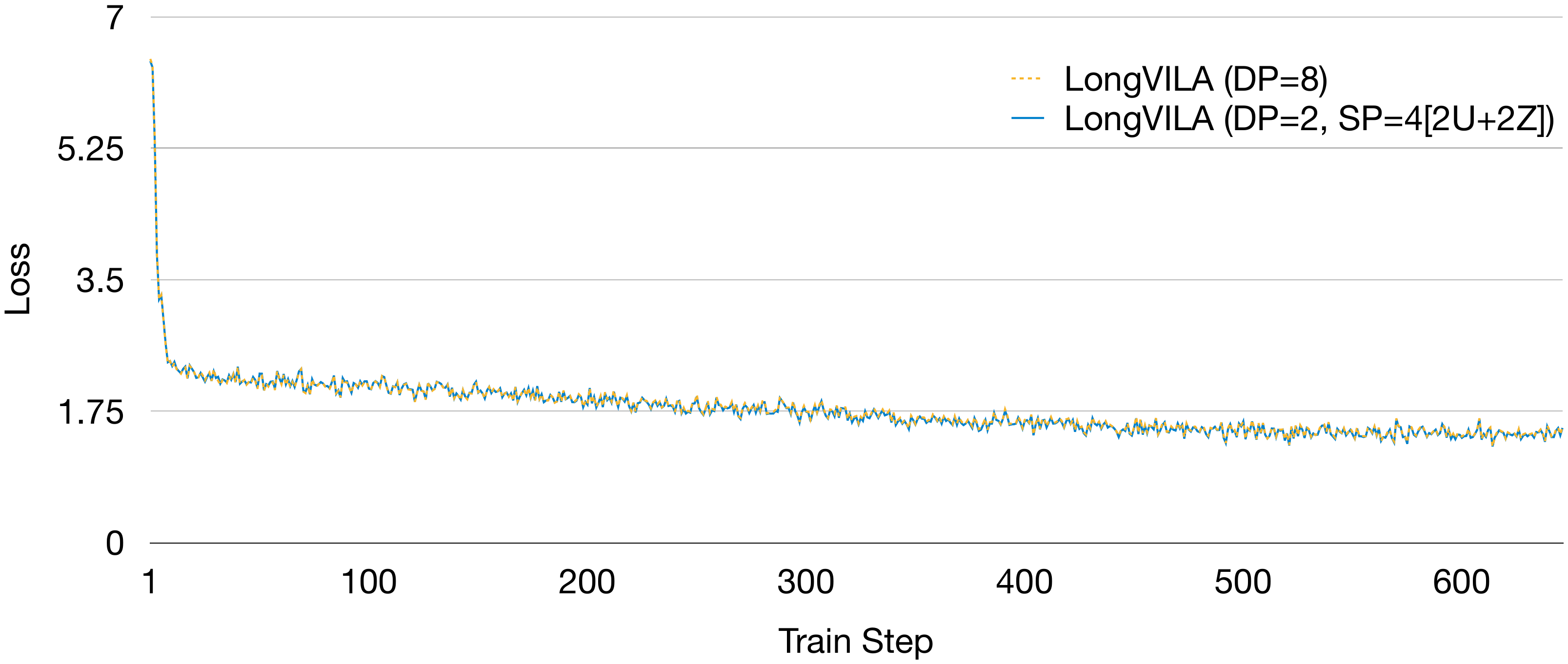}
    \caption{Convergence evaluation. We compare the training loss curves for LongVILA with and without sequence parallelism. This figure illustrates the convergence of the LongVILA model on 8 H100 GPUs, with a sequence parallelism degree of 4, compared to pure data parallelism training. 2U+2Z indicates that 2D-Attention mechanism is enabled, where both the Ulysses and Zigzag-Ringttn degrees are 2. The training dataset is \texttt{shot2story}. The two curves align closely, indicating that our MM-SP system does not negatively impact training quality.}
    \label{fig:convergence}
\end{figure}

\begin{table}[h]
\centering
\caption{\dacheng{Our training system, used in conjunction with FSDP~\citep{zhao2023pytorch} or Zero-3~\citep{rajbhandari2020zero}, on 32 H100 GPUs. We found that FSDP offers more efficient memory management, which led us to select it as our default configuration. (Time per iteration, seconds).}}
\resizebox{\columnwidth}{!}{
\color{DLTODO}\begin{tabular}{l|c|c|c|c|c|c}
\toprule
\multirow{2}{*}{\begin{tabular}[c]{@{}l@{}}Sequence\\ Length\end{tabular}}  & \multicolumn{3}{c}{Zero-3} & \multicolumn{3}{c}{FSDP} \\
 & \zigzag & Ulysses & 2D Attention & \zigzag & Ulysses & 2D Attention\\
\midrule
320 K & OOM & OOM & OOM & 23.57 & 10.70 & 11.12  \\
288 K & OOM & OOM & OOM & 20.24 & 8.68 & 8.65 \\
256 K & OOM & OOM & OOM & 17.54 & 6.98 & 7.04 \\
224 K & 19.04 & 7.06 & 5.73 & 15.22 & 5.47 & 5.53 \\
192 K & 13.01 & 4.24 & 4.38 & 12.97 & 4.15 & 4.24 \\
160 K & 10.73 & 3.09 & 3.23 & 10.83 & 3.02 & 3.11 \\
128 K & 8.63 & 2.16 & 2.30 & 8.38 & 2.07 & 2.17 \\
96 K & 6.49 & 1.43 & 1.53 & 6.35 & 1.33 & 1.41 \\
64 K & 4.40 & 1.01 & 1.08 & 4.25 & 0.76 & 0.80 \\
32 K & 2.06 & 1.58 & 1.04 & 2.26 & 0.39 & 0.40 \\
\bottomrule
\end{tabular}}
\label{tab:training_system_efficiency-appendix}
\end{table}

\begin{table}[h]
\centering
\caption{\dacheng{Training system throughput comparison on 64 H100 GPUs, measured in time per iteration (seconds). Ulysses is not included in this comparison as it supports only up to 32 GPUs.}}
\color{DLTODO}\begin{tabular}{l|c|c|c|c|c}
\toprule
Sequence length & \multicolumn{2}{|c|}{Megatron-LM} & \multicolumn{3}{|c}{Ours} \\
\midrule & CP & CP=8+TP=8 & \zigzag & Ulysses & 2D Attention\\
\midrule
640 K & OOM & OOM & 88.4 & - & OOM \\
578 K & OOM & OOM & 77.2 & - & 16.9 \\
512 K & OOM & OOM & 66.1 & - & 13.31 \\
448 K & OOM & OOM & 57.5 & - & 10.39 \\
384 K & OOM & OOM & 48.6 & - & 7.80 \\
320 K & OOM & OOM & 40.5 & - & 5.63 \\
256 K & OOM & 5.31 & 32.2 & - & 3.93 \\
192 K & 8.81 & 3.10 & 24.1 & - & 2.49 \\
128 K & 7.10 & 1.57 & 16.0 & - & 1.36 \\
64 K & 3.09 & 0.61 & 8.04 & - & 0.57 \\
32 K & 1.86 & 0.44 & 4.24 & - & 0.33 \\
\bottomrule
\end{tabular}
\label{tab:training_system_efficiency_64GPUs}
\end{table}

\begin{table*}[t]
\centering
\caption{
Comparison with state-of-the-art methods~\citep{li2023blip, Dai2023InstructBLIP, bai2023qwen, liu2023improved, lin2023vila, liu2024llavanext, tong2024cambrian1fullyopenvisioncentric, li2024minigeminiminingpotentialmultimodality} on 10 image based VLM benchmarks. S3 refers to the stage 3 model in LongVILA training pipeline.}
\resizebox{\columnwidth}{!}{
\begin{tabular}{l|l|l| lllll | lllll }
\toprule
Method & LLM & Res. & VQA$^\text{v2}$ & GQA & VizWiz &  SQA$^\text{I}$ & VQA$^\text{T}$& MMB & MMB$^\text{CN}$ & SEED & LLaVA$^\text{W}$ & MM-Vet \\
\midrule

BLIP-2 & Vicuna-13B & 224 & 41.0 & 41 & 19.6 & 61 & 42.5 &  -- & -- & 46.4 & 38.1 & 22.4 \\ \hline
 \multirow{2}{*}{InstructBLIP} & Vicuna-7B & 224& -- & 49.2 & 34.5 & 60.5 & 50.1  & 36 & 23.7 & 53.4 & 60.9 & 26.2 \\
 & Vicuna-13B & 224 & -- & 49.5 & 33.4 & 63.1 & 50.7 & -- & -- & -- & 58.2 & 25.6 \\ \hline
Qwen-VL & Qwen-7B & 448 & 78.8 & 59.3 & 35.2& 67.1 & 63.8 & 38.2 & 7.4 & 56.3 & -- & -- \\
Qwen-VL-Chat & Qwen-7B & 448 & 78.2 & 57.5 & 38.9 & 68.2 & 61.5 & 60.6 & 56.7 & 58.2 & -- & -- \\
 \hline
 \multirow{2}{*}{LLaVA-1.5} & Vicuna-1.5-7B & 336 & 78.5 & 62.0 & 50.0 & 66.8 & 58.2 & 64.3 & 58.3 & 58.6 & 63.4 & 30.5 \\ 
 & Vicuna-1.5-13B & 336& 80.0 & 63.3 & 53.6 & 71.6 & 61.3 & 67.7 & 63.6 & 61.6 & 70.7 & 35.4 \\
 \hline
 \multirow{2}{*}{VILA} & Llama 2-7B & 336& 79.9 & 62.3 & 57.8 & 68.2 & 64.4& 68.9 & 61.7 & 61.1 & 69.7 & 34.9 \\
 & Llama 2-13B & 336& \underline{80.8} & 63.3 & \underline{60.6} & 73.7 & 66.6 & 70.3 & \underline{64.3} & \underline{62.8} & 73.0 & \underline{38.8} \\
 \hline
LLaVA-NeXT-8B & Llama 3-8B & 672 & -- & \underline{65.2} & -- & 72.8 & 64.6 & 72.1 & -- & -- & \underline{80.1} & -- \\
Cambrian-1-8B & Llama 3-8B & 1024 & -- & 64.6 & -- & \underline{80.4} & \underline{71.7} & \underline{75.9} & -- & -- & -- & -- \\
Mini-Gemini-HD-8B & Llama 3-8B & 1536 & -- & 64.5 & -- & 75.1 & 70.2 & 72.7 & -- & -- & -- & -- \\
\midrule
LongVILA-7B (S3) & Qwen2-7B & dynamic & \textbf{85.4} & \textbf{65.4} & \textbf{65.0} & \textbf{98.5} & \textbf{77.8} & \textbf{83.4} & \textbf{80.0} & \textbf{70.6} & \textbf{77.6} & \textbf{51.7} \\
\bottomrule
\end{tabular}
}
\label{tab:sota_comparison}
\end{table*}

\begin{table*}[t]
\centering
\caption{{
Detailed model complexity analysis of LongVILA among various model size, number of frames, parameters, latency (ms), and TFLOPs. We profile LongVILA models into 4 types of components. Image Encoder includes vision tower and mm projector. LLM Linears include k/q/v/o projectors and linears in decoder layers. LLM Attention is the attention computation. LLM Others include other components, like embeddings, output heads, and normalization layers. We use fp16 data type, Flash-Attention2~\citep{flash-attention2} on one A100 GPU for latency measurement. As the number of frames increases, the FLOPs and latency of LLM Attention grow quadratically, whereas other components increase linearly. LLM Attention represents the predominant computational cost in long video understanding, highlighting the MM-SP system.}}
\resizebox{\linewidth}{!}{
{
\begin{tabular}{l|l|l|cccc|cccc}
\toprule
\multirow{3}{*}{Frames} & \multirow{3}{*}{Context} & \multirow{2}{*}{Metric} & \multicolumn{4}{c|}{LongVILA-1.5B}                                                                                                                                                                                               & \multicolumn{4}{c}{LongVILA-7B}                                                                                                                                                                                                  \\
                        &                          &                         & \begin{tabular}[c]{@{}c@{}}Image\\ Encoder\end{tabular} & \begin{tabular}[c]{@{}c@{}}LLM\\ Linears\end{tabular} & \begin{tabular}[c]{@{}c@{}}LLM\\ Attention\end{tabular} & \begin{tabular}[c]{@{}c@{}}LLM\\ Others\end{tabular} & \begin{tabular}[c]{@{}c@{}}Image\\ Encoder\end{tabular} & \begin{tabular}[c]{@{}c@{}}LLM\\ Linears\end{tabular} & \begin{tabular}[c]{@{}c@{}}LLM\\ Attention\end{tabular} & \begin{tabular}[c]{@{}c@{}}LLM\\ Others\end{tabular} \\ \midrule
\multicolumn{1}{c}{}                         & \multicolumn{1}{c}{}                          & \multicolumn{1}{l|}{Params}                   & 0.44B                               & 1.31B        & -             & 0.23B                            & 0.46B          & 6.53B        &    -           & 1.09B       \\ \midrule
\multirow{2}{*}{32}     & \multirow{2}{*}{6415}    & Latency                 & 196.0                                                   & 109.3                                                 & 27.0                                                    & 28.9                                                 & 201.0                                                   & 426.2                                                 & 51.1                                                    & 42.1                                                 \\
                        &                          & TFLOPs                  & 5.13                                                    & 4.20                                                  & 1.77                                                    & 0.75                                                 & 5.22                                                    & 20.93                                                 & 4.13                                                    & 1.74                                                 \\ \midrule
\multirow{2}{*}{64}     & \multirow{2}{*}{12719}   & Latency                 & 375.2                                                   & 208.1                                                 & 80.7                                                    & 50.9                                                 & 375.8                                                   & 831.1                                                 & 173.8                                                   & 80.8                                                 \\
                        &                          & TFLOPs                  & 10.26                                                   & 8.33                                                  & 6.96                                                    & 1.48                                                 & 10.45                                                   & 41.50                                                 & 16.24                                                   & 3.46                                                 \\ \midrule
\multirow{2}{*}{128}    & \multirow{2}{*}{25327}   & Latency                 & 740.4                                                   & 403.2                                                 & 288.1                                                   & 97.1                                                 & 755.1                                                   & 1642.4                                                & 644.9                                                   & 158.3                                                \\
                        &                          & TFLOPs                  & 20.52                                                   & 16.60                                                 & 27.59                                                   & 2.95                                                 & 20.89                                                   & 82.64                                                 & 64.38                                                   & 6.89                                                 \\ \midrule
\multirow{2}{*}{256}    & \multirow{2}{*}{50543}   & Latency                 & 1456.0                                                  & 811.2                                                 & 1087.7                                                  & 192.5                                                & 1476.3                                                  & 3308.2                                                & 2529.6                                                  & 325.1                                                \\
                        &                          & TFLOPs                  & 41.05                                                   & 33.12                                                 & 109.89                                                  & 5.88                                                 & 41.78                                                   & 164.92                                                & 256.40                                                  & 13.74                                                \\ \midrule
\multirow{2}{*}{512}    & \multirow{2}{*}{100975}  & Latency                 & 2921.0                                                  & 1660.9                                                & 4359.2                                                  & 389.4                                                & 2980.2                                                  & 6675.1                                                & 10149.3                                                 & 653.4                                                \\
                        &                          & TFLOPs                  & 82.09                                                   & 66.17                                                 & 438.59                                                  & 11.75                                                & 83.57                                                   & 329.48                                                & 1023.37                                                 & 27.45                                                \\ \bottomrule
\end{tabular}}
}
\label{tab:latency_flops_profile}
\end{table*}

\end{document}